\documentclass[iicol,sn-mathphys]{sn-jnl}% Default with double column layout

\jyear{2025}%

\usepackage{graphicx}%
\usepackage{multirow}%
\usepackage{multicol}
\usepackage{xurl}
\usepackage{xcolor}
\usepackage{placeins}
\usepackage{array}
\usepackage{framed}
\usepackage{balance}
\usepackage{needspace}
\usepackage{adjustbox}
\usepackage{microtype}
\newcommand{\pixel}{\mathbf{q}}

\newcommand{\vecsymbol}[1]{\mathbf{#1}}

\makeatletter
\newcommand{\stdcline}[1]{\@cline#1\@nil}% LaTeX's own \cline (sn-jnl turns \cline into \cmidrule)
\makeatother

\theoremstyle{thmstyleone}%
\theoremstyle{thmstyletwo}%

\theoremstyle{thmstylethree}%

\begin{document}

\title[Article Title]{Geometric Prior-Guided Neural Implicit Surface Reconstruction in the Wild}

\author[1]{\fnm{Lintao} \sur{Xiang}}\email{ltxiang.work@gmail.com}

\author[1]{\fnm{Hongpei} \sur{Zheng}}\email{hongpei.zheng@manchester.ac.uk}

\author*[2]{\fnm{Bailin} \sur{Deng}}\email{DengB3@cardiff.ac.uk}

\author[1]{\fnm{Hujun} \sur{Yin}}\email{hujun.yin@manchester.ac.uk}

\affil[1]{\orgdiv{Department of Electrical and Electronic Engineering}, \orgname{The University of Manchester}, \orgaddress{\street{Oxford Road}, \city{Manchester}, \postcode{M13 9PL}, \country{UK}}}

\affil*[2]{\orgdiv{School of Computational and Mathematical Sciences}, \orgname{Cardiff University}, \orgaddress{\street{Senghennydd Road}, \city{Cardiff}, \postcode{CF24 4AG}, \country{UK}}}

\abstract{Neural implicit surface reconstruction using volume rendering techniques has recently achieved significant advancements in creating high-fidelity surfaces from multiple 2D images. However, current methods primarily target scenes with consistent illumination and struggle to accurately reconstruct 3D geometry in uncontrolled environments with transient occlusions or varying appearances. While some neural radiance field (NeRF)-based variants can better manage photometric variations and transient objects in complex scenes, they are designed for novel view synthesis and lack the surface constraints needed for precise surface reconstruction. To overcome this limitation, we introduce a novel approach that applies multiple geometric constraints to the implicit surface optimization process, enabling more accurate reconstructions from unconstrained image collections. First, we utilize sparse 3D points from structure-from-motion (SfM) to refine the signed distance function estimation for the reconstructed surface, with a displacement compensation to accommodate noise in the sparse points. Additionally, we employ robust normal priors derived from a normal predictor, filtered using edge detection and a reprojection consistency check, to improve alignment with the actual surface geometry. Experiments show that our method achieves the best average performance among the compared methods on the Heritage-Recon benchmark of internet photo collections and performs comparably to the strongest baselines on two standard multi-view datasets. Our approach enables high-quality 3D reconstruction of landmarks and objects, making it applicable to diverse scenarios such as digital preservation of cultural heritage sites and 3D object modeling.}


\maketitle

\section{Introduction}\label{sec1}

\begin{figure*}[t]
\centering
\includegraphics[width=\textwidth]{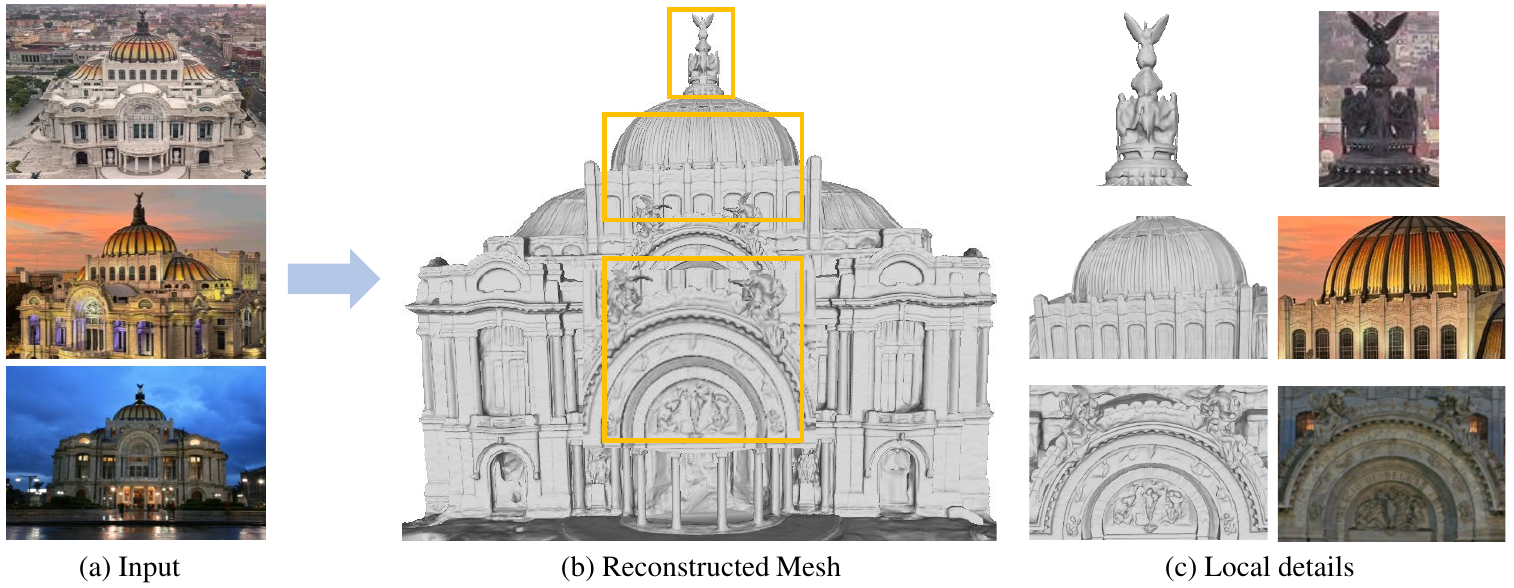}
\caption{The proposed neural 3D surface reconstruction method aims to reconstruct high-fidelity 3D surfaces from unstructured internet photos of landmarks with varying appearances and complex occlusions. (a) Examples of input images of the \textit{Palacio de Bellas Artes}. (b) Reconstructed 3D mesh of the landmark. (c) Close-ups of the regions marked in (b), shown with the corresponding regions of the input images.}
\label{fig:teaser}
\end{figure*}

3D surface reconstruction seeks to capture the shape and appearance of real objects or scenes from multi-view images. It integrates techniques from computer vision and graphics to analyze and visualize complex data from various media sources, enhancing digital content creation and interactive experiences. Traditional multi-view stereo (MVS) methods~\cite{sch2016pw, cernea2020openmvs} typically follow a multi-step process that includes depth map estimation, point cloud fusion, and surface reconstruction. This process often results in the accumulation of errors at each step. In response, recent advances have introduced neural implicit surface reconstruction techniques~\cite{yariv2020multiview,wang2021neus,darmon2022improving,wang2022hf, li2022vox, petrov2023anise} to reconstruct detailed meshes for both objects and scenes. These methods employ neural networks to implicitly represent surfaces and utilize neural volume rendering to enhance the reconstruction process. Notably, they adopt the signed distance function (SDF) for surface representation and apply the SDF-based density function to refine the learning of the implicit surface.
Like neural radiance fields (NeRF)~\cite{mildenhall2021nerf} and many of its extensions (e.g.,~\cite{xu2022point}), these methods assume a static scene under constant illumination. In contrast, internet photos are taken by different people at different times, so illumination and appearance vary between images (see Fig.~\ref{fig:teaser}(a)), and transient objects such as pedestrians often occlude the scene.
Consequently, applying these methods directly to such photos may result in significant performance degradation and lead to reconstructions that exhibit severe artifacts and excessive smoothing. To address these complex scenarios, various NeRF variants~\cite{martin2021nerf,chen2022hallucinated} have been developed to learn controllable appearance and mitigate occlusions by segmenting scenes into static and transient components. While effective at generating novel view images, these approaches generally struggle to create high-fidelity meshes for complex, uncontrolled settings or internet photo collections. NeuralRecon-W~\cite{sun2022neural} is the first to introduce a hybrid voxel- and surface-guided sampling technique to improve reconstruction from unstructured internet data, accommodating variations in lighting and occlusions. However, its accuracy is limited by the absence of explicit geometric constraints.

In this paper, we propose GeoNeucon-W, which incorporates geometric priors into NeuralRecon-W to recover more accurate and detailed surfaces from internet photos (Fig.~\ref{fig:teaser}(b) and (c)). We employ two types of priors: sparse 3D points derived from structure from motion (SfM)~\cite{schonberger2016structure} and monocular normal priors. Both priors have been used in previous work; for example, Geo-NeuS~\cite{fu2022geo} uses sparse points, NeuRIS~\cite{wang2022neuris} uses normal priors, and Ea-NeuS~\cite{li2023edge} uses both for indoor scenes. However, these priors can be inaccurate, and our contributions mainly lie in how we handle such inaccuracies. For sparse points, Geo-NeuS and Ea-NeuS directly penalize the SDF values at the SfM points, which may pull the surface toward misplaced points. In contrast, we project each point toward the current surface along the SDF gradient and penalize the SDF value at the projected location. This displacement compensation serves as a regularizer on the SDF rather than forcing the surface to pass through each point (Sec.~\ref{sec:point_comp}), and our experiments show that it is less sensitive to Gaussian perturbations of the points than direct supervision.

Monocular normal priors~\cite{wang2022neuris,zehao2022monosdf} provide useful constraints on surface orientation, but not all of them are reliable. NeuRIS selects normal priors by checking the photometric consistency of image patches across views, while Ea-NeuS observes that normal priors are unreliable in regions with complex shapes or thin structures and discards those located at image edges. We adopt this edge criterion and further employ a reprojection consistency check, which is commonly used to filter MVS depth maps~\cite{yao2018mvsnet}, to remove the normal priors at points occluded in adjacent views.

We evaluate GeoNeucon-W on Heritage-Recon~\cite{sun2022neural}, a benchmark of internet photo collections, and on two standard multi-view datasets, Tanks and Temples~\cite{Knapitsch2017} and DTU~\cite{jensen2014large}. On Heritage-Recon, our method achieves the best average performance among the compared methods, including NeuralRecon-W. On Tanks and Temples and DTU, it performs comparably to the strongest baselines.
It also recovers finer geometric details than NeuralRecon-W on internet photos of three additional landmarks. Ablations on two Heritage-Recon scenes show that each component improves the reconstruction.

Our main contributions can be summarized as follows:
\begin{itemize}
\item We propose a displacement compensation regularizer, which improves the reconstruction by evaluating the SDF at SfM points projected toward the current surface instead of the original points.

\item We combine edge-based rejection with a reprojection consistency check to filter monocular normal priors, with each filter improving the reconstruction.

\item We show that our method achieves the best average performance on Heritage-Recon and is competitive on two other datasets, and analyze each component through ablations and a Gaussian point-perturbation study.
\end{itemize}

\section{Related Work}\label{sec2}

Surface reconstruction is a fundamental area in computer graphics and computer vision, focusing on creating detailed and precise surface models from limited or imperfect input data. In this section, we review previous work, spanning from classical to contemporary approaches, with an emphasis on studies that are most closely aligned with our research.

\subsection{Multi-View Reconstruction}
Multi-view Stereo (MVS) aims to reconstruct 3D scene geometry from multiple images with corresponding camera parameters. Traditional MVS methods~\cite{goesele2006multi, furukawa2009accurate, sch2016pw} first estimate per-view depth maps and then fuse the depth maps into a dense point cloud. Afterward, surface reconstruction algorithms such as Poisson reconstruction~\cite{kazhdan2006poisson} or Delaunay triangulation~\cite{labatut2009robust} are applied to reconstruct 3D meshes. Recently, learning-based MVS methods~\cite{yao2018mvsnet,gu2020cascade,yang2020cost,ding2022transmvsnet, xiang2024feature} have made remarkable advances over traditional methods. However, these approaches encounter problems related to complex computations and global inconsistencies caused by a redundant pipeline involving depth estimation, depth filtering, point cloud fusion and surface reconstruction.

\subsection{Neural Implicit Surface Reconstruction}
Surface reconstruction algorithms can be broadly categorized into explicit  and implicit methods. Explicit methods represent surfaces using discrete structures like voxel grids~\cite{broadhurst2001probabilistic, seitz1999photorealistic} or meshes~\cite{boissonnat1993three,labatut2009robust}. These methods are limited in accuracy by their resolution, where finer details are lost at lower resolutions. Implicit methods represent surfaces continuously with an implicit function. This allows for the extraction of surfaces at any desired resolution.

Recently, neural implicit scene representations have demonstrated impressive performance. NeRF~\cite{mildenhall2021nerf} pioneered this approach by representing a scene as a neural radiance field, using a Multi-Layer Perceptron (MLP) to synthesize novel views. Since then, numerous improvements have been proposed to optimize scene representations. While NeRF-based methods have excelled at rendering high-quality images, they often struggle with high-fidelity surface reconstruction. To address this challenge, methods like VolSDF~\cite{yariv2021volume} and NeuS~\cite{wang2021neus} have introduced SDF-based weight functions that can simultaneously optimize the color and geometric properties of a scene. This enables them to achieve complete surface reconstruction without requiring additional mask information. 
Further refinements have been proposed by {enforcing multi-view photometric consistency~\cite{darmon2022improving,fu2022geo}, incorporating monocular depth and normal priors~\cite{zehao2022monosdf,wang2022neuris} or depth measurements from RGB-D sensors~\cite{azinovic2022neural,zhu2022nice}, or modeling high-frequency details in the SDF~\cite{wang2022hf},} to enhance the accuracy of implicit functions and capture finer geometric details.

\subsection{Reconstruction in the Wild}
Most NeRF-based and NeuS-based methods are designed to reconstruct scenes or objects under controlled settings and struggle with unstructured, variable, and complex scenes, such as internet photo collections. To address these challenges, NeRF-W~\cite{martin2021nerf} and Ha-NeRF~\cite{chen2022hallucinated} incorporate appearance embedding and a transient head to manage variable illumination and occlusions, respectively. Other approaches tackle object occlusions using external priors such as segmentation information~\cite{kobayashi2022decomposing, lee2023semantic} or masks~\cite{wang2022dm, yang2023cross}.

However, these NeRF variants primarily focus on novel view synthesis from unconstrained images and generally fail to produce high-quality 3D meshes. Sun et al.~\cite{sun2022neural} combine the benefits of NeRF-W and NeuS into a method that efficiently reconstructs surface geometry for in-the-wild scenes. This approach introduces a hybrid voxel- and surface-guided sampling method that significantly reduces training time and sets a new benchmark with corresponding metrics and evaluation protocols for unstructured internet photos. Nevertheless, this method does not reconstruct detailed surface features due to a lack of geometric constraints. Besides, Neuralangelo \cite{li2023neuralangelo} leverages numerical gradient regularization and a coarse-to-fine training strategy to boost surface reconstruction quality. However, it still struggles with the challenging scenarios of in-the-wild internet photos.

\begin{figure*}[t]
\centering
    \includegraphics[width=\textwidth]{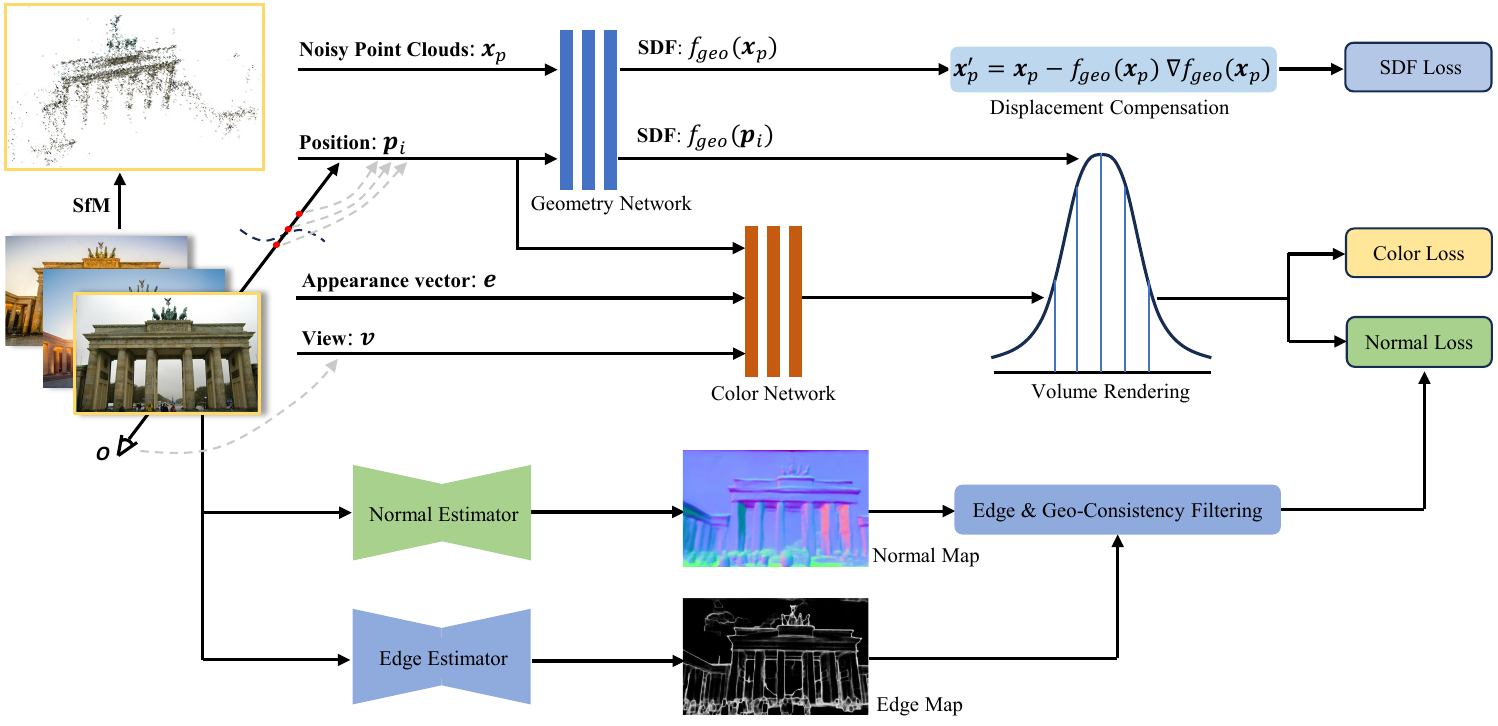}
    \caption{The pipeline of the proposed method. Sparse points from COLMAP~\cite{schonberger2016structure} are projected toward the current surface using the SDF and its gradient, and the SDF values at the projected locations are penalized (displacement compensation). 
    Normal priors from a pre-trained estimator supervise the surface normals through the normal loss. Priors at image edges are discarded, and during training, a reprojection consistency check based on the current SDF discards priors at points occluded in adjacent views.}
   \label{fig:framework}
\end{figure*}

Recently, 3D Gaussian Splatting (3DGS)~\cite{kerbl20233d} has attracted attention for its explicit Gaussian representation and real-time novel view synthesis. WildGaussians~\cite{kulhanek2024wildgaussians}, SWAG~\cite{dahmani2024swag}, and GS-W~\cite{zhang2024gsw} extend 3DGS to unconstrained photo collections by modeling appearance variation and transient objects. DroneSplat~\cite{tang2025dronesplat} handles dynamic distractors in in-the-wild drone imagery. GaRe~\cite{bai2025gare} enables relighting of outdoor scenes from unconstrained photo collections. However, these methods mainly target novel view synthesis or relighting rather than accurate surface reconstruction. Gaussian representations have also been adapted for surface reconstruction, with SuGaR~\cite{guedon2024sugar} aligning Gaussians with the surface for mesh extraction and 2DGS~\cite{huang2024_2dgs} using oriented planar Gaussian disks with geometric regularization. These methods, however, are designed for images captured under consistent conditions. In contrast, our method builds on an SDF representation and focuses on how sparse-point and normal priors can be used to reconstruct surfaces from internet photos.

\section{Methods}\label{sec3}
\label{sec:method}

Given unconstrained internet photos of a scene, GeoNeucon-W represents its geometry with an SDF learned through volume rendering, following NeuralRecon-W~\cite{sun2022neural}. Since color supervision alone can leave geometric ambiguities~\cite{fu2022geo}, we further constrain the SDF with two geometric priors. Specifically, we use sparse SfM points to regularize the SDF through displacement compensation, which penalizes the SDF value at each point after projecting it toward the current surface. The projection is computed from the current SDF and its gradient, so no additional network is required. As discussed in Sec.~\ref{sec:point_comp}, this term acts as a regularizer rather than forcing the surface to pass through the points. Additionally, we supervise the surface orientation with monocular normal priors, which are filtered by the edge criterion of Ea-NeuS~\cite{li2023edge} and a reprojection consistency check. Fig.~\ref{fig:framework} summarizes the pipeline, and the following subsections describe each component.

\subsection{Geometry and Color Networks}
\label{sec:preliminary}
To accurately reconstruct surface geometry, we represent the scene using a geometry function $f_{geo}$ and a color function $f_{color}$, similar to~\cite{sun2022neural}.
Each function is represented using a multi-layer perceptron (MLP).
The geometry function takes a spatial point $\vecsymbol{x} \in \mathbb{R}^3$ as input and produces the signed distance from  $\vecsymbol{x}$ to the reconstructed surface $S$, so that $S$ is the zero level-set of   $f_{geo}$:
\begin{equation}
S = \{ \vecsymbol{x} \in \mathbb{R}^3 \mid f_{geo}( \vecsymbol{x})=0 \}.
\end{equation} 
The input to the color function $f_{color}$ includes a spatial point $\vecsymbol{x}$, a view direction $\vecsymbol{v} \in \mathbb{S}^2$, and the appearance embedding vector $\vecsymbol{e}$ of an input image that models its lighting condition and is optimized during training. The output of $f_{color}$ is the emitted color $\mathbf{c}$ at $\vecsymbol{x}$ according to the view direction and the lighting condition, i.e., $\mathbf{c} = f_{color}(\vecsymbol{x}, \vecsymbol{v}, \vecsymbol{e})$.  
The color function is utilized to compute the predicted colors for the pixels in the input images through volume rendering, which are used to optimize the implicit representation with the supervision from their ground-truth colors. Specifically, given a pixel $\pixel{}$, 
we emit a ray from $\pixel{}$ and derive a set of points on the ray using sample parameters $\{t_j \geq 0 \}_{j = 1}^n$ as
$ \vecsymbol{x}(t_j)= \vecsymbol{o}+t_j \vecsymbol{v}$,
where $\mathbf{o}$ is the camera center, and $\mathbf{v}$ is the ray direction. The predicted color $\hat{C}(\pixel{})$ is then computed by accumulating the emitted colors of the sample points via:
\begin{equation}
\begin{aligned}
\hat{C}(\pixel) &=\sum_{j=1}^{n} T_j\alpha_j f_{color}(\vecsymbol{x}(t_j),\vecsymbol{v},\vecsymbol{e}), \\
T_j &=\prod_{k=1}^{j-1} (1-\alpha_k),
\end{aligned}
\label{eq:ColorAccumulation}
\end{equation}
where $T_j$ is the accumulated transmittance, and $\alpha_j$ is the discrete opacity defined as
\begin{equation}
\alpha_j = \max \left(\frac{\Phi_s(u_j)-\Phi_s(u_{j+1})}{\Phi_s(u_j)},0\right),
\end{equation}
where $u_j=f_{geo}(\vecsymbol{x}(t_j))$, and $\Phi_s$ is the sigmoid function with a learnable parameter $s$~\cite{wang2021neus}.

\begin{figure}[t]
\centering
    \includegraphics[width=\columnwidth]{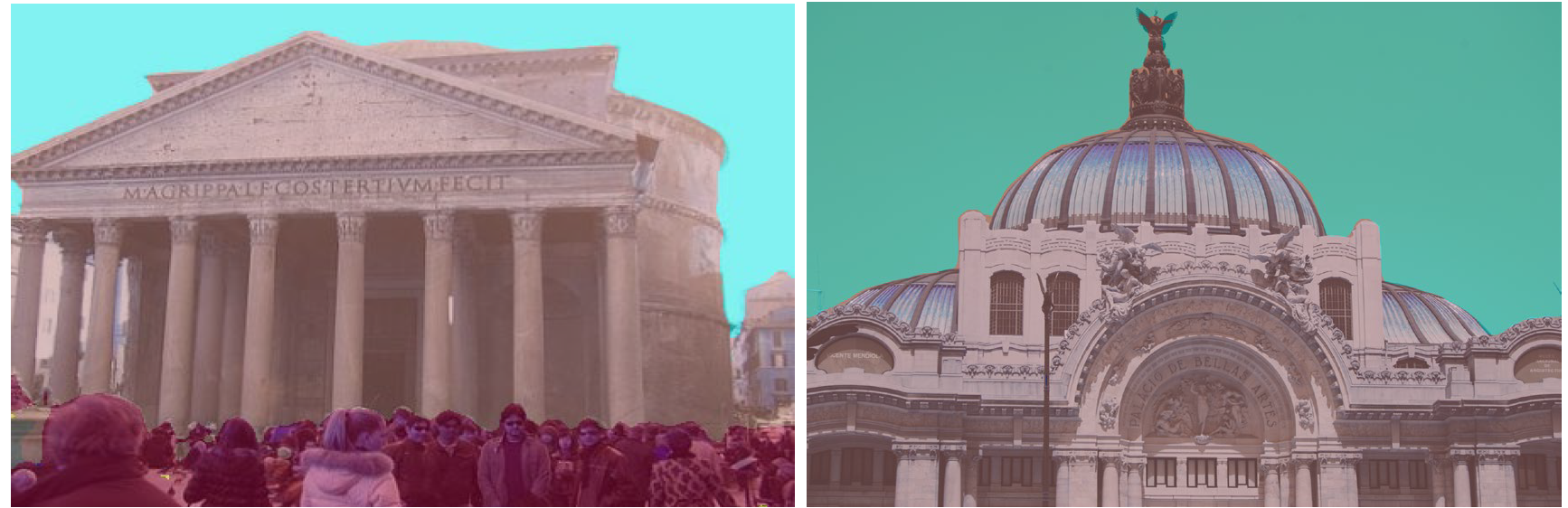}
    \caption{Examples of semantic masks on \textit{Pantheon Exterior} (left) and \textit{Palacio de Bellas Artes} (right). The masks are used to exclude transient objects, such as pedestrians, from training and to identify the sky for the mask loss.}
   \label{fig:seg}

\end{figure}

To handle the occlusions in internet photo collections, we use the semantic segmentation algorithm toolbox within OpenMMLab~\cite{mmseg2020} to identify occluded areas (see Fig.~\ref{fig:seg} for an example) in each input image, and exclude the pixels in these areas from training. Let $\mathcal{Q}$ be the set of remaining pixels from all input images. Then we introduce a color loss for these pixels to penalize the difference between the predicted color $\hat{C}(\pixel)$ and the ground-truth color ${C}(\pixel)$ for each pixel $\pixel$ in  $\mathcal{Q}$:
\begin{equation}
\mathcal{L}_{color} = \frac{1}{\lvert\mathcal{Q}\rvert} \sum_{\pixel \in \mathcal{Q}} \|C(\pixel) - \hat{C}(\pixel)\|.
\end{equation}

In addition, since in-the-wild scenes may contain the textureless sky in the background, which can cause spherical shells in the extracted meshes, we follow~\cite{sun2022neural} and introduce a mask loss $\mathcal L_{mask}$ to remove the sky with the help of semantic masks as shown in Fig.~\ref{fig:seg}, making the foreground region more complete:
\begin{equation}
\mathcal L_{mask} = \mathrm{BCE}(\hat{W},1-M^{bp}),
\end{equation} 
where $\mathrm{BCE}$ is the binary cross entropy loss, $\hat{W}$ is a map defined on the images and computed using the sum of weights along the camera ray for each pixel, and $M^{bp}$ is the background mask obtained from OpenMMLab \cite{mmseg2020}, which is 1 for background pixels and 0 otherwise.

\subsection{Geometric Prior-Guided Optimization}

Although the color network presented in Sec.~\ref{sec:preliminary} accommodates varying lighting conditions, color fitting alone does not determine accurate geometry. We therefore incorporate two geometric priors, namely sparse SfM points for displacement compensation and filtered monocular normal priors for orientation supervision, to guide the optimization and recover finer surface details. These priors are described in the next two subsections.

\subsubsection{Sparse Points Compensation}\label{sec:point_comp}

Color supervision through volume rendering does not directly constrain the surface-ray intersections~\cite{fu2022geo}. Sparse SfM points can provide such constraints, because an accurate SDF should be zero at any point lying on the true surface. Geo-NeuS~\cite{fu2022geo} and Ea-NeuS~\cite{li2023edge} therefore penalize the SDF values directly at the observed SfM points.

However, SfM points may deviate from the true surface because of matching and triangulation errors, and direct supervision can then pull the surface toward misplaced points. Ea-NeuS reduces this effect by discarding points with few feature matches, but it still penalizes the SDF at the observed coordinates of the remaining points. To make the supervision less sensitive to such errors, we instead evaluate the SDF at displaced locations computed from the SDF and its gradient. For an exact SDF, the gradient has unit norm near a regular part of the zero level-set, and the step
\begin{equation}
\label{eq:displace}
\vecsymbol{x}'_p = \vecsymbol{x}_p - f_{geo}(\vecsymbol{x}_p) \nabla f_{geo}(\vecsymbol{x}_p)
\end{equation}
projects a point $\vecsymbol{x}_p$ onto the zero level-set. We therefore define the displacement compensation loss as the SDF residual at the displaced locations:
\begin{equation}
\label{eq:sdf_loss}
\mathcal{L}_{sdf} = \frac{1}{\lvert\mathcal{P}\rvert}\sum_{\vecsymbol{x}_p \in \mathcal{P}}\lvert{f_{geo}}(\vecsymbol{x}'_p) \rvert,
\end{equation} 
where $\mathcal{P}$ is the sparse point set generated by SfM. Since Eq.~\eqref{eq:displace} projects points onto the zero level-set of an exact SDF, $\mathcal{L}_{sdf}$ vanishes for any such SDF, regardless of where its surface lies relative to the SfM points. For example, if $f_{geo}$ is the exact SDF of a plane, $\mathcal{L}_{sdf}$ is zero wherever the plane is placed. The loss is thus nonzero only where $f_{geo}$ deviates from an exact SDF near these points, and it does not by itself keep the surface close to them. We therefore use $\mathcal{L}_{sdf}$ as a regularizer on the SDF within the total loss, with the SfM points determining where the regularization is applied rather than where the surface must lie. Similar to~\cite{wang2021neus}, we also introduce a loss $\mathcal{L}_{eik}$ that encourages $f_{geo}$ to satisfy the Eikonal equation $\|\nabla f_{geo}\| = 1$, which holds for an exact SDF:
\begin{equation} 
\mathcal{L}_{eik} = 
\frac{1}{\sum_{\pixel \in \mathcal{Q}} \lvert\mathcal{T}_{\pixel}\rvert} \sum_{\pixel \in \mathcal{Q}} \sum_{\vecsymbol{x} \in \mathcal{T}_{\pixel}} (\|\nabla f_{geo}(\vecsymbol{x}) \| - 1)^2,
\end{equation} 
where $\mathcal{T}_{\pixel}$ is the set of sample points along the ray for the pixel $\pixel$ according to Eq.~\eqref{eq:ColorAccumulation}.

\subsubsection{Robust Normal Priors} 
\label{eq:RobustNormalPriors}
The point-based regularizer discussed above uses sparse spatial samples but provides no direct surface-orientation supervision. To further improve surface quality, we also utilize surface normals as priors to constrain the surface shape. Specifically, it is well known in differential geometry that the surface normal at a point determines the direction of its tangent plane. Together with the point position, it provides a first-order approximation of the surface around the point, which results in a stronger prior than the point position alone.
Therefore, we constrain the implicit surface normals with robust normal priors derived from the input images during reconstruction.

To this end, we first use the pre-trained normal predictor from~\cite{wang2022neuris} to produce a unit normal $\vecsymbol{n}_c(\pixel)$ in camera coordinates for each pixel ${\pixel}$ in the set $\mathcal{Q}$, and transform it into the world frame of $f_{geo}$ to obtain the normal prior $\vecsymbol{n}_p(\pixel) = R_{cw}^{\top}\vecsymbol{n}_c(\pixel)$, where $R_{cw}$ is the world-to-camera rotation of the image containing $\pixel$.
Let $\vecsymbol{x}_{\pixel}^*$ denote the intersection point between the ray emitted from $\pixel$ and the zero level-set of $f_{geo}$. We then compute the unit surface normal $\hat{\vecsymbol{n}}({\pixel})$ at $\vecsymbol{x}_{\pixel}^*$ via
\begin{equation}
\hat{\vecsymbol{n}}({\pixel}) = %\frac
{\nabla f_{geo}(\vecsymbol{x}_{\pixel}^*)}\big/{\|\nabla f_{geo}(\vecsymbol{x}_{\pixel}^*)\|},
\end{equation}
and enforce the consistency between the surface normal $\hat{\vecsymbol{n}}(\pixel)$ and the prior $\vecsymbol{n}_{p}(\pixel)$.
To determine the intersection point $\vecsymbol{x}_{\pixel}^*$, we search for all neighboring sample point pairs $(\vecsymbol{x}(t_j), \vecsymbol{x}(t_k))$ on the ray such that $f_{geo}(\vecsymbol{x}(t_j))$ and $f_{geo}(\vecsymbol{x}(t_k))$ are of opposite signs. For each such pair, we estimate the intersection point $\vecsymbol{x}_{jk}$ between the zero level-set of $f_{geo}$ and the segment $\overline{\vecsymbol{x}(t_j), \vecsymbol{x}(t_k)}$ via:
\begin{equation}
\vecsymbol{x}_{jk}=\frac{f_{geo}(\vecsymbol{x}(t_j)) \cdot \vecsymbol{x}(t_k) - f_{geo}( \vecsymbol{x}(t_k)) \cdot \vecsymbol{x}(t_j)}{f_{geo}( \vecsymbol{x}(t_j))-f_{geo}( \vecsymbol{x}(t_k))}.
\end{equation} 
Then among all such intersection points, we choose the one closest to the camera center as $\vecsymbol{x}_{\pixel}^*$.

It should be noted, however, that not all normal priors are reliable. Therefore, we introduce two criteria to filter out the unreliable normal priors. 
First, as observed in Ea-NeuS~\cite{li2023edge}, normal priors are not reliable in areas characterized by complex shapes or thin structures. Following Ea-NeuS, we leverage a pre-trained edge detection network from~\cite{su2021pixel} to identify such areas, marking pixels whose predicted edge probability exceeds an empirically chosen threshold $\tau_e = 0.7$, and remove their normal priors.

\begin{figure}[t]
\centering
    \includegraphics[width=\columnwidth]{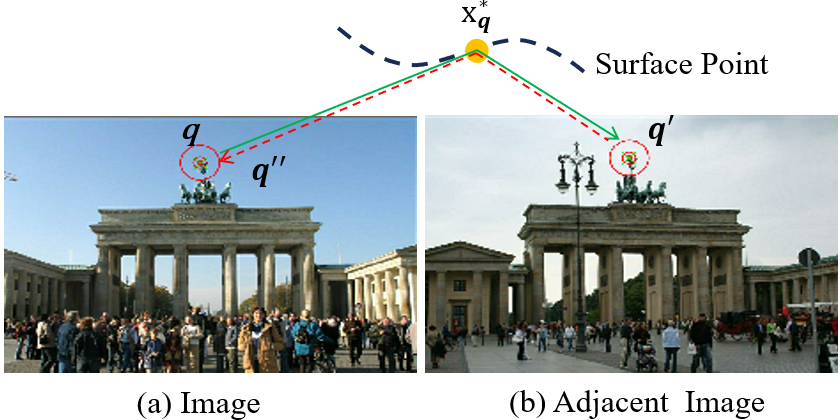}
    \caption{Reprojection consistency check. A pixel $\pixel$ is warped to an adjacent image through its surface point $\vecsymbol{x}_{\pixel}^*$ to obtain $\pixel'$, which is then warped back to obtain $\pixel''$. If $\|\pixel-\pixel''\| \geq \tau$, the point is likely occluded in the adjacent image, and the normal prior at $\pixel$ is removed.}
   \label{fig:consis}
\end{figure}

In addition, for a point that is visible in one image but occluded in an adjacent image, the normal prior may also be unreliable and should be removed. We identify such points during training using a reprojection consistency check based on the current SDF. Let $\pixel$ be a pixel from an image $I_1$, $\vecsymbol{x}_{\pixel}^*$ be the estimated intersection point between its ray and the surface as explained above, and $I_2$ be an adjacent image. We first warp $\pixel$ from $I_1$ to $I_2$ by projecting the point $\vecsymbol{x}_{\pixel}^*$ onto the image plane of $I_2$ to obtain a pixel $\pixel'$. Then we warp $\pixel'$ back from $I_2$ to $I_1$ to obtain another pixel $\pixel''$ (see Fig.~\ref{fig:consis}). If the reprojection error $\|\pixel - \pixel''\| \geq \tau$ (with $\tau{=}1$ pixel by default), we consider the normal prior at $\pixel$ to be unreliable and exclude it from the training process.
A sensitivity study of $\tau$ is reported in Sec.~\ref{sec:reproj_sens}.

Finally, for the set of remaining pixels $\mathcal{Q}_{n}$ with reliable normal priors after the above filtering, we introduce a normal loss term for surface optimization:
\begin{equation}
\mathcal{L}_{normal} = \frac{1}{\lvert\mathcal{Q}_{n}\rvert} \sum_{\pixel \in \mathcal{Q}_n} (1- \vecsymbol{{n}}_p(\pixel) \cdot \hat{\vecsymbol{n}}(\pixel)) .
\end{equation}

\subsection{Loss Function}

\label{sec:loss}
Combining the above loss terms, we obtain the total loss for the proposed method as:
\begin{align}
\mathcal{L}_{{total}} =\, & \lambda_1 \mathcal{L}_{color}
+ \lambda_2 \mathcal{L}_{normal} \notag \\
& + \lambda_3 \mathcal{L}_{sdf}
+ \lambda_4 \mathcal{L}_{eik}
+ \lambda_5 \mathcal{L}_{mask}.
\end{align}
We set $\lambda_1=1.0$, $\lambda_2=1.0$, $\lambda_3=1.0$, $\lambda_4=0.01$, and $\lambda_5=0.1$ in our experiments. We fix the weight of the color loss as the reference and choose the other weights empirically. %Sec.~\ref{sec:lambda_sens} analyzes their sensitivity.}

\section{Experiments}\label{sec4}

\subsection{Implementation Details}

\begin{table*}[t]
\caption{Quantitative results on the Heritage-Recon dataset~\cite{sun2022neural}. For COLMAP, we include two variants with different values of the octree depth $d$. The best results are in \textbf{bold}, and the second-best are \underline{underlined}. The proposed method achieves the best performance in average F1 score among the compared methods. Low, Medium, and High denote the scene-specific distance thresholds in Table~\ref{tab:error}, with Low the strictest.}
\centering
\let\cline\stdcline% plain \cline in this table: no extra space, joins the vertical rules
\renewcommand{\arraystretch}{1.2}% the same spacing for every row
\resizebox{\textwidth}{!}
{
\setlength{\tabcolsep}{4.5pt}
\begin{tabular}{ll|lll lll lll  lll l }
\hline
& \multicolumn{1}{l|} {\multirow{2}{*}{\textbf{Method}}} & \multicolumn{3}{c|}{\!\!\!\!\!\!\!\!\! \textbf{BG}}  & \multicolumn{3}{c|}{\!\!\!\!\!\!\!\!\! \textbf{LM}} & \multicolumn{3}{c|}{\!\!\!\!\!\!\!\textbf{PE}} & \multicolumn{3}{c|}{\!\!\!\!\!\!\!\textbf{PBA}} & \multicolumn{1}{l} {\multirow{2}{*}{\textbf{Mean F1 $\uparrow$ }}} \\ 

\cline{3-14} 

\multirow{11}{*}{ \textbf{Low}} & 
\multicolumn{1}{l|}{}                        
& \multicolumn{1}{l}{P $\uparrow$}  & \multicolumn{1}{l}{R $\uparrow$}  & \multicolumn{1}{l|}{F1 $\uparrow$} 
& \multicolumn{1}{l}{P $\uparrow$}  & \multicolumn{1}{l}{R $\uparrow$}  & \multicolumn{1}{l|}{F1 $\uparrow$}  
& \multicolumn{1}{l}{P $\uparrow$}  & \multicolumn{1}{l}{R $\uparrow$}  & \multicolumn{1}{l|}{F1 $\uparrow$}
& \multicolumn{1}{l}{P $\uparrow$}  & \multicolumn{1}{l}{R $\uparrow$}  & \multicolumn{1}{l|}{F1 $\uparrow$}
\\ \hline

& COLMAP ($d$=11) \cite{sch2016pw}
& \multicolumn{1}{l}{61.2} & \multicolumn{1}{l}{40.6} & \multicolumn{1}{l|}{48.8}
& \multicolumn{1}{l}{20.4} & \multicolumn{1}{l}{23.6} & \multicolumn{1}{l|}{21.9}
& \multicolumn{1}{l}{48.7} & \multicolumn{1}{l}{27.8} & \multicolumn{1}{l|}{35.4}
& \multicolumn{1}{l}{\textbf{91.6}} & \multicolumn{1}{l}{49.7} & \multicolumn{1}{l|}{64.5}
& \multicolumn{1}{c}{42.7}
\\ 
& COLMAP ($d$=13)  \cite{sch2016pw}     
& \multicolumn{1}{l}{\textbf{79.1}} & \multicolumn{1}{l}{41.6} & \multicolumn{1}{l|}{\underline{54.5}}  
& \multicolumn{1}{l}{25.8} & \multicolumn{1}{l}{32.6} & \multicolumn{1}{l|}{28.8}  
& \multicolumn{1}{l}{\underline{54.4}} & \multicolumn{1}{l}{\textbf{71.0}} & \multicolumn{1}{l|}{\textbf{61.6}}  
& \multicolumn{1}{l}{75.5} & \multicolumn{1}{l}{28.9} & \multicolumn{1}{l|}{41.8}  
& \multicolumn{1}{c}{46.7} 
\\ 

\cline{2-15}

& VisMVSNet \cite{zhang2020visibility}    
& \multicolumn{1}{l}{48.2} & \multicolumn{1}{l}{6.6} & \multicolumn{1}{l|}{11.6}
& \multicolumn{1}{l}{23.4} & \multicolumn{1}{l}{27.0} & \multicolumn{1}{l|}{25.1}
& \multicolumn{1}{l}{36.1} & \multicolumn{1}{l}{22.9} & \multicolumn{1}{l|}{28.1}
& \multicolumn{1}{l}{\underline{81.0}} & \multicolumn{1}{l}{42.5} & \multicolumn{1}{l|}{55.7} 
& \multicolumn{1}{c}{30.1}
\\ 
%61.2 47.6 53.6; 
& GeoMVSNet \cite{zhang2023geomvsnet}    
& \multicolumn{1}{l}{61.2} & \multicolumn{1}{l}{47.6} & \multicolumn{1}{l|}{53.6}
& \multicolumn{1}{l}{30.8} & \multicolumn{1}{l}{32.8} & \multicolumn{1}{l|}{31.7}
& \multicolumn{1}{l}{46.5} & \multicolumn{1}{l}{48.8} & \multicolumn{1}{l|}{47.6}
& \multicolumn{1}{l}{77.1} & \multicolumn{1}{l}{55.7} & \multicolumn{1}{l|}{64.7} 
& \multicolumn{1}{c}{49.4}
\\ 
\cline{2-15}
& NeRF-W \cite{martin2021nerf}                          
& \multicolumn{1}{l}{1.8} & \multicolumn{1}{l}{1.1} & \multicolumn{1}{l|}{1.4} 
& \multicolumn{1}{l}{28.0} & \multicolumn{1}{l}{13.4} & \multicolumn{1}{l|}{18.1}
& \multicolumn{1}{l}{11.0} & \multicolumn{1}{l}{3.4} & \multicolumn{1}{l|}{5.2}
& \multicolumn{1}{l}{39.6} & \multicolumn{1}{l}{12.7} & \multicolumn{1}{l|}{19.2}
& \multicolumn{1}{c}{11.0}
\\ %62.9 45.4 52.8 
& Geo-NeuS \cite{fu2022geo}                          
& \multicolumn{1}{l}{62.9} & \multicolumn{1}{l}{45.4} & \multicolumn{1}{l|}{52.8} 
& \multicolumn{1}{l}{30.5} & \multicolumn{1}{l}{32.1} & \multicolumn{1}{l|}{31.3}
& \multicolumn{1}{l}{46.3} & \multicolumn{1}{l}{48.2} & \multicolumn{1}{l|}{47.2}
& \multicolumn{1}{l}{76.9} & \multicolumn{1}{l}{55.3} & \multicolumn{1}{l|}{64.3}
& \multicolumn{1}{c}{48.9}
\\ %61.4 48.3 54.1 
& Neuralangelo \cite{li2023neuralangelo}                    
& \multicolumn{1}{l}{61.4} & \multicolumn{1}{l}{\underline{48.3}} & \multicolumn{1}{l|}{54.1} 
& \multicolumn{1}{l}{31.8} & \multicolumn{1}{l}{34.7} & \multicolumn{1}{l|}{33.2}
& \multicolumn{1}{l}{47.2} & \multicolumn{1}{l}{49.5} & \multicolumn{1}{l|}{48.3}
& \multicolumn{1}{l}{77.7} & \multicolumn{1}{l}{56.2} & \multicolumn{1}{l|}{65.2}
& \multicolumn{1}{c}{50.2}
\\ 

& NeuralRecon-W \cite{sun2022neural}                         
& \multicolumn{1}{l}{62.6} & \multicolumn{1}{l}{47.9} & \multicolumn{1}{l|}{54.3} 
& \multicolumn{1}{l}{\underline{32.8}} & \multicolumn{1}{l}{\underline{35.6}} & \multicolumn{1}{l|}{\underline{34.1}} 
& \multicolumn{1}{l}{48.5} & \multicolumn{1}{l}{50.8} & \multicolumn{1}{l|}{49.6} 
& \multicolumn{1}{l}{78.1} & \multicolumn{1}{l}{\underline{57.7}} & \multicolumn{1}{l|}{\underline{66.4}} 
& \multicolumn{1}{c}{\underline{51.1}} 
\\
& Ours                                  
& \multicolumn{1}{l}{\underline{65.2}} & \multicolumn{1}{l}{\textbf{50.4}} & \multicolumn{1}{l|}{\textbf{56.8}}
& \multicolumn{1}{l}{\textbf{34.8}} & \multicolumn{1}{l}{\textbf{36.7}} & \multicolumn{1}{l|}{\textbf{35.7}}
& \multicolumn{1}{l}{\textbf{55.6}} & \multicolumn{1}{l}{\underline{53.8}} & \multicolumn{1}{l|}{\underline{54.7}}
& \multicolumn{1}{l}{79.9} & \multicolumn{1}{l}{\textbf{58.5}} & \multicolumn{1}{l|}{\textbf{67.5}}
& \multicolumn{1}{c}{\textbf{53.7}}
\\  \hline

\multirow{9}{*}{ \textbf{Medium}} 
& COLMAP ($d$=11) \cite{sch2016pw}                       & \multicolumn{1}{l}{73.7} & \multicolumn{1}{l}{53.1} & \multicolumn{1}{l|}{61.8}
& \multicolumn{1}{l}{50.1} & \multicolumn{1}{l}{55.1} & \multicolumn{1}{l|}{52.4}
& \multicolumn{1}{l}{66.6} & \multicolumn{1}{l}{56.5} & \multicolumn{1}{l|}{61.2}
& \multicolumn{1}{l}{\textbf{98.3}} & \multicolumn{1}{l}{57.9} & \multicolumn{1}{l|}{72.9}
 & \multicolumn{1}{c}{62.1}
\\ 
& COLMAP ($d$=13) \cite{sch2016pw}      
& \multicolumn{1}{l}{\textbf{89.3}} & \multicolumn{1}{l}{53.1} & \multicolumn{1}{l|}{66.6}  
& \multicolumn{1}{l}{57.7} & \multicolumn{1}{l}{60.6} & \multicolumn{1}{l|}{59.1}  
& \multicolumn{1}{l}{\underline{74.8}} & \multicolumn{1}{l}{\textbf{79.7}} & \multicolumn{1}{l|}{\textbf{77.1}}  
& \multicolumn{1}{l}{86.8} & \multicolumn{1}{l}{\textbf{71.5}} & \multicolumn{1}{l|}{\underline{78.4}}  
& \multicolumn{1}{c}{70.3}  \\ 
    \cline{2-15}
& VisMVSNet \cite{zhang2020visibility}  
& \multicolumn{1}{l}{72.4} & \multicolumn{1}{l}{12.1} & \multicolumn{1}{l|}{20.8}
& \multicolumn{1}{l}{46.3} & \multicolumn{1}{l}{42.7} & \multicolumn{1}{l|}{44.4}
& \multicolumn{1}{l}{57.8} & \multicolumn{1}{l}{34.4} & \multicolumn{1}{l|}{43.1}
& \multicolumn{1}{l}{\underline{95.1}} & \multicolumn{1}{l}{57.8} & \multicolumn{1}{l|}{71.9}
& \multicolumn{1}{c}{45.1}
\\ %77.8 63.5 69.9
& GeoMVSNet \cite{zhang2023geomvsnet}  
& \multicolumn{1}{l}{77.8} & \multicolumn{1}{l}{63.5} & \multicolumn{1}{l|}{69.9}
& \multicolumn{1}{l}{65.6} & \multicolumn{1}{l}{66.9} & \multicolumn{1}{l|}{66.2}
& \multicolumn{1}{l}{70.2} & \multicolumn{1}{l}{69.8} & \multicolumn{1}{l|}{69.9}
& \multicolumn{1}{l}{86.6} & \multicolumn{1}{l}{{67.9}} & \multicolumn{1}{l|}{76.1}
& \multicolumn{1}{c}{70.5}
\\ \cline{2-15}

& NeRF-W \cite{martin2021nerf}                          
& \multicolumn{1}{l}{4.1} & \multicolumn{1}{l}{5.0} & \multicolumn{1}{l|}{4.5} 
& \multicolumn{1}{l}{51.9} & \multicolumn{1}{l}{24.8} & \multicolumn{1}{l|}{33.6}
& \multicolumn{1}{l}{21.9} & \multicolumn{1}{l}{6.2} & \multicolumn{1}{l|}{9.7}
& \multicolumn{1}{l}{64.3} & \multicolumn{1}{l}{28.8} & \multicolumn{1}{l|}{39.8}
& \multicolumn{1}{c}{21.9}
\\ %76.7 62.8 69.1 
& Geo-NeuS \cite{fu2022geo}                          
& \multicolumn{1}{l}{76.7} & \multicolumn{1}{l}{62.8} & \multicolumn{1}{l|}{69.1} 
& \multicolumn{1}{l}{65.3} & \multicolumn{1}{l}{66.2} & \multicolumn{1}{l|}{65.7}
& \multicolumn{1}{l}{69.8} & \multicolumn{1}{l}{69.3} & \multicolumn{1}{l|}{69.5}
& \multicolumn{1}{l}{86.1} & \multicolumn{1}{l}{67.5} & \multicolumn{1}{l|}{75.6}
& \multicolumn{1}{c}{70.0}
\\  %79.0 64.8 71.2 
& Neuralangelo \cite{li2023neuralangelo}                          
& \multicolumn{1}{l}{79.0} & \multicolumn{1}{l}{64.8} & \multicolumn{1}{l|}{71.2} 
& \multicolumn{1}{l}{66.2} & \multicolumn{1}{l}{67.1} & \multicolumn{1}{l|}{66.6}
& \multicolumn{1}{l}{70.8} & \multicolumn{1}{l}{70.3} & \multicolumn{1}{l|}{70.5}
& \multicolumn{1}{l}{87.5} & \multicolumn{1}{l}{68.4} & \multicolumn{1}{l|}{76.7}
& \multicolumn{1}{c}{71.3}
\\  
& NeuralRecon-W \cite{sun2022neural}
& \multicolumn{1}{l}{78.7} & \multicolumn{1}{l}{\underline{65.7}} & \multicolumn{1}{l|}{\underline{71.6}} 
& \multicolumn{1}{l}{\underline{67.7}} & \multicolumn{1}{l}{\underline{68.7}} & \multicolumn{1}{l|}{\underline{68.2}} 
& \multicolumn{1}{l}{71.7} & \multicolumn{1}{l}{71.1} & \multicolumn{1}{l|}{71.4} 
& \multicolumn{1}{l}{88.3} & \multicolumn{1}{l}{69.7} & \multicolumn{1}{l|}{77.9} 
 & \multicolumn{1}{c}{\underline{72.3}} 
\\
& Ours                                  
& \multicolumn{1}{l}{\underline{79.9}} & \multicolumn{1}{l}{\textbf{66.4}} & \multicolumn{1}{l|}{\textbf{72.5}}
& \multicolumn{1}{l}{\textbf{68.1}} & \multicolumn{1}{l}{\textbf{69.3}} & \multicolumn{1}{l|}{\textbf{68.7}}
& \multicolumn{1}{l}{\textbf{75.3}} & \multicolumn{1}{l}{\underline{74.9}} & \multicolumn{1}{l|}{\underline{75.1}}
& \multicolumn{1}{l}{89.6} & \multicolumn{1}{l}{\underline{70.1}} & \multicolumn{1}{l|}{\textbf{78.6}}

& \multicolumn{1}{c}{\textbf{73.7}}
\\  \hline
\multirow{9}{*}{ \textbf{High}} 
& COLMAP ($d$=11) \cite{sch2016pw}      
& \multicolumn{1}{l}{81.4} & \multicolumn{1}{l}{59.0} & \multicolumn{1}{l|}{68.4}
& \multicolumn{1}{l}{70.7} & \multicolumn{1}{l}{74.0} & \multicolumn{1}{l|}{72.3}
& \multicolumn{1}{l}{73.8} & \multicolumn{1}{l}{69.4} & \multicolumn{1}{l|}{71.5}
& \multicolumn{1}{l}{\textbf{99.2}} & \multicolumn{1}{l}{63.7} & \multicolumn{1}{l|}{77.6}
& \multicolumn{1}{c}{72.5}
\\ 
& COLMAP ($d$=13) \cite{sch2016pw}   
& \multicolumn{1}{l}{\textbf{94.1}} & \multicolumn{1}{l}{58.8} & \multicolumn{1}{l|}{72.4}  
& \multicolumn{1}{l}{78.0} & \multicolumn{1}{l}{75.9} & \multicolumn{1}{l|}{76.9}  
& \multicolumn{1}{l}{\textbf{81.6}} & \multicolumn{1}{l}{\textbf{83.6}} & \multicolumn{1}{l|}{\textbf{82.6}}  
& \multicolumn{1}{l}{91.3} & \multicolumn{1}{l}{\textbf{80.8}} & \multicolumn{1}{l|}{\underline{85.7}}  
& \multicolumn{1}{c}{79.4}  
\\ \cline{2-15}

& VisMVSNet \cite{zhang2020visibility}   
& \multicolumn{1}{l}{85.6} & \multicolumn{1}{l}{14.5} & \multicolumn{1}{l|}{24.8}
& \multicolumn{1}{l}{63.8} & \multicolumn{1}{l}{53.0} & \multicolumn{1}{l|}{57.9}
& \multicolumn{1}{l}{68.2} & \multicolumn{1}{l}{42.3} & \multicolumn{1}{l|}{52.2}
& \multicolumn{1}{l}{\underline{98.0}} & \multicolumn{1}{l}{64.4} & \multicolumn{1}{l|}{77.7}
& \multicolumn{1}{c}{{53.2}}
\\ %83.1 72.1 77.2
& GeoMVSNet \cite{zhang2023geomvsnet}   
& \multicolumn{1}{l}{83.1} & \multicolumn{1}{l}{72.1} & \multicolumn{1}{l|}{77.2}
& \multicolumn{1}{l}{78.9} & \multicolumn{1}{l}{79.4} & \multicolumn{1}{l|}{79.1}
& \multicolumn{1}{l}{78.6} & \multicolumn{1}{l}{76.8} & \multicolumn{1}{l|}{77.6}
& \multicolumn{1}{l}{91.2} & \multicolumn{1}{l}{75.4} & \multicolumn{1}{l|}{82.5}
 & \multicolumn{1}{c}{79.1}
\\ \cline{2-15}

& NeRF-W \cite{martin2021nerf}             
& \multicolumn{1}{l}{7.6} & \multicolumn{1}{l}{11.5} & \multicolumn{1}{l|}{9.2} 
& \multicolumn{1}{l}{67.3} & \multicolumn{1}{l}{34.2} & \multicolumn{1}{l|}{45.3}
& \multicolumn{1}{l}{28.8} & \multicolumn{1}{l}{8.3} & \multicolumn{1}{l|}{12.9}
& \multicolumn{1}{l}{79.5} & \multicolumn{1}{l}{40.5} & \multicolumn{1}{l|}{53.7}
& \multicolumn{1}{c}{30.3}
\\ %82.6 71.3 76.5
& Geo-NeuS \cite{fu2022geo}             
& \multicolumn{1}{l}{82.6} & \multicolumn{1}{l}{71.3} & \multicolumn{1}{l|}{76.5} 
& \multicolumn{1}{l}{78.5} & \multicolumn{1}{l}{79.2} & \multicolumn{1}{l|}{78.8}
& \multicolumn{1}{l}{78.2} & \multicolumn{1}{l}{76.5} & \multicolumn{1}{l|}{77.3}
& \multicolumn{1}{l}{90.9} & \multicolumn{1}{l}{74.8} & \multicolumn{1}{l|}{82.1}
& \multicolumn{1}{c}{78.7}
\\ %84.8 72.8 78.4
& Neuralangelo \cite{li2023neuralangelo}             
& \multicolumn{1}{l}{84.8} & \multicolumn{1}{l}{72.8} & \multicolumn{1}{l|}{78.4} 
& \multicolumn{1}{l}{81.0} & \multicolumn{1}{l}{81.2} & \multicolumn{1}{l|}{81.1}
& \multicolumn{1}{l}{79.1} & \multicolumn{1}{l}{77.2} & \multicolumn{1}{l|}{78.1}
& \multicolumn{1}{l}{92.1} & \multicolumn{1}{l}{76.2} & \multicolumn{1}{l|}{83.3}
& \multicolumn{1}{c}{80.2}
\\ 
& NeuralRecon-W \cite{sun2022neural}     
& \multicolumn{1}{l}{85.9} & \multicolumn{1}{l}{\underline{73.7}} & \multicolumn{1}{l|}{\underline{79.3}} 
& \multicolumn{1}{l}{\underline{82.1}} & \multicolumn{1}{l}{\underline{82.1}} & \multicolumn{1}{l|}{\underline{82.1}} 
& \multicolumn{1}{l}{79.9} & \multicolumn{1}{l}{77.9} & \multicolumn{1}{l|}{78.9} 
& \multicolumn{1}{l}{93.0} & \multicolumn{1}{l}{77.1} & \multicolumn{1}{l|}{84.3} 
& \multicolumn{1}{c}{\underline{81.2}}

\\
& Ours                                  
& \multicolumn{1}{l}{\underline{87.4}} & \multicolumn{1}{l}{\textbf{74.9}} & \multicolumn{1}{l|}{\textbf{80.7}}
& \multicolumn{1}{l}{\textbf{84.6}} & \multicolumn{1}{l}{\textbf{84.7}} & \multicolumn{1}{l|}{\textbf{84.6}}
& \multicolumn{1}{l}{\underline{81.2}} & \multicolumn{1}{l}{\underline{82.9}} & \multicolumn{1}{l|}{\underline{82.0}}
& \multicolumn{1}{l}{94.5} & \multicolumn{1}{l}{\underline{78.8}} & \multicolumn{1}{l|}{\textbf{85.9}}
& \multicolumn{1}{c}{\textbf{83.3}}
\\  \hline

\end{tabular}
}
\label{tab:overall}
\end{table*}

Following NeuralRecon-W~\cite{sun2022neural}, for the geometry network $f_{geo}$, we use an 8-layer MLP with 512 hidden units, and for the color network $f_{color}$, we use a 4-layer MLP with 256 hidden units. We use the hybrid voxel- and surface-guided sampling strategy proposed in~\cite{sun2022neural} for volume rendering, with both the number of voxel-guided samples $n_v$ and the number of surface-guided samples $n_s$ set to 8. The proposed model is trained for 20 epochs on four NVIDIA A100 GPUs, with a calculated training time of $10.19$\,h on \textit{BG}. After training, we extract 3D meshes from the SDF represented by $f_{geo}$ using the Marching Cubes algorithm~\cite{lorensen1998marching}.
Our source code is provided in the supplementary materials.

Normal priors are generated with the pre-trained monocular normal estimator used in the preprocessing pipeline of NeuRIS~\cite{wang2022neuris}, and edge priors with PiDiNet~\cite{su2021pixel} using its \texttt{table5\_pidinet} checkpoint. Both networks are used with their pre-trained weights, without fine-tuning on any of our datasets. Their input images are resized so that the longer side is 640 pixels while preserving the aspect ratio, and the predicted maps are bilinearly resampled to the training image resolution when necessary. 
%PiDiNet outputs continuous edge probabilities. We form a binary mask $E(\pixel)=\mathbf{1}[p_e(\pixel)\geq\tau_e]$ using an empirically set probability threshold $\tau_e=0.7$, where $p_e(\pixel)$ is the predicted edge probability. Normal priors at pixels with $E(\pixel)=1$ are discarded before reprojection consistency checking. The dimensionless probability threshold $\tau_e$ is distinct from the reprojection threshold $\tau=1$ pixel. The predicted normal and edge maps are computed once before training and cached. The reprojection consistency check is performed during training using the current SDF, as described in Sec.~3.2.2. 
The normal and edge maps are computed once before training and cached, whereas the reprojection consistency check is performed during training with the current SDF (Sec.~\ref{eq:RobustNormalPriors}).
The computational costs are reported in Sec.~\ref{sec:cost}.

\begin{figure*}[t]
\centering
    \includegraphics[width=\textwidth]{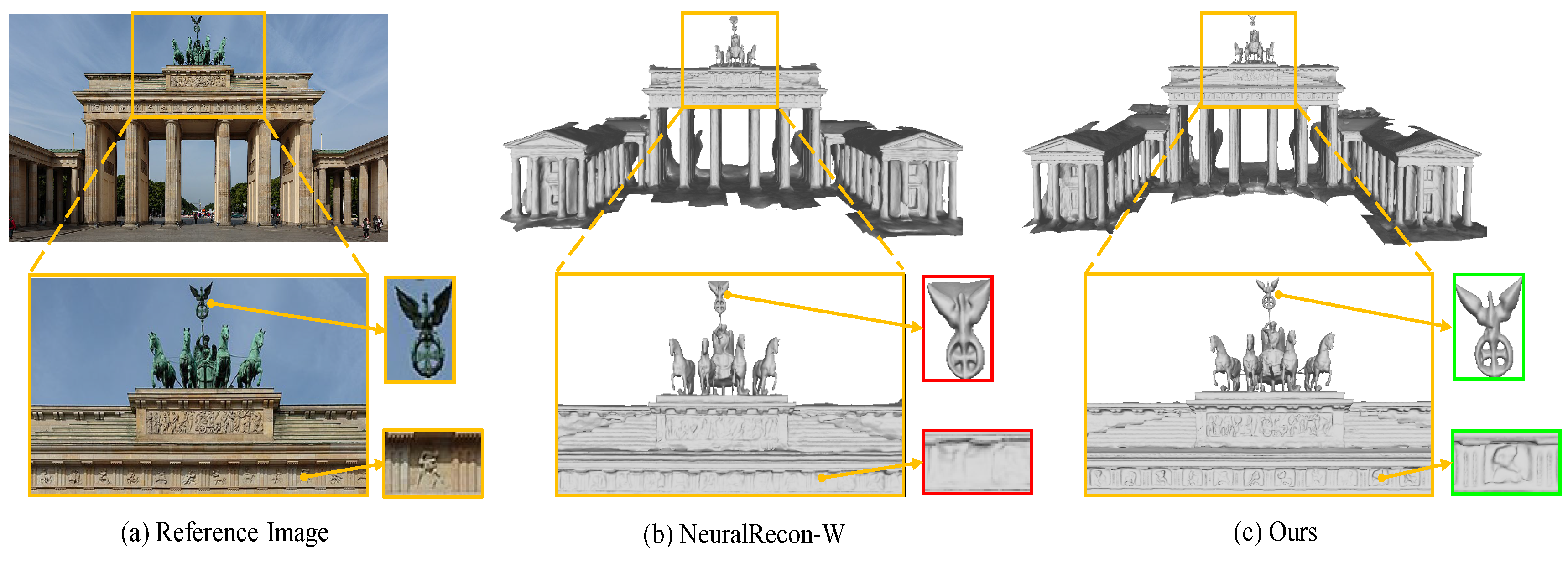}
    \caption{
    Qualitative comparison between our method and NeuralRecon-W~\cite{sun2022neural} on \textit{BG}. As shown in the reconstructed meshes and corresponding local regions, the proposed method can better capture fine details. Close-ups are outlined in red for NeuralRecon-W and in green for our method. The F1 score at the Low threshold increases from 54.3 for NeuralRecon-W to 56.8 for our method (Table~\ref{tab:overall}).}
  \label{fig:bg}
\end{figure*}

\begin{figure*}[t]
\centering
    \includegraphics[width=1.0\textwidth]{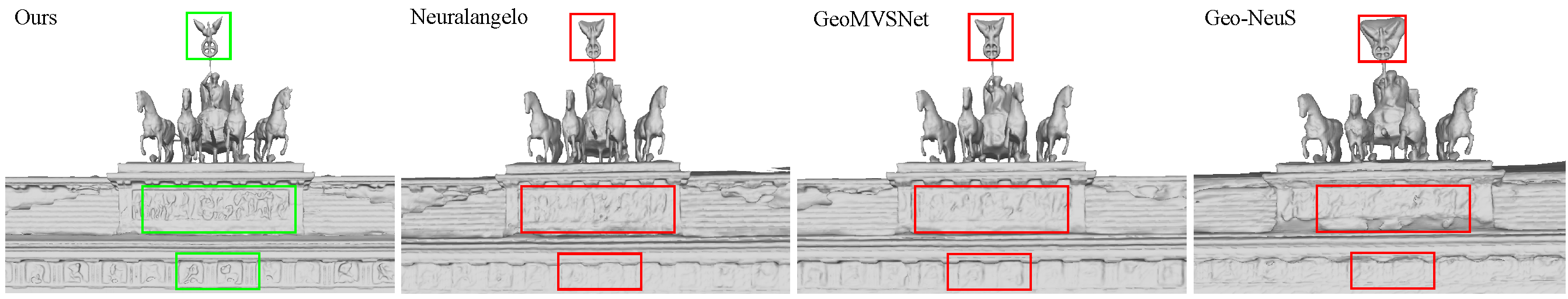}
  \caption{Qualitative comparison with more baselines on \textit{BG}. Our method more accurately captures scene details than other approaches, which tend to produce incomplete or noisy surfaces. The highlighted regions are outlined in green for our method and in red for the baselines. The F1 scores at the Low threshold are 56.8 for our method, 54.1 for Neuralangelo, 53.6 for GeoMVSNet, and 52.8 for Geo-NeuS (Table~\ref{tab:overall}).}
   \label{fig:baseline_bg_fig}
\end{figure*}

\subsection{Evaluation on Heritage-Recon Dataset}
We evaluated the performance of GeoNeucon-W using the benchmark proposed in~\cite{sun2022neural} with their
Heritage-Recon dataset. The dataset consists of internet photo collections of several well-known landmarks with ground-truth shapes. It contains four cultural heritage sites: \textit{Brandenburg Gate} (BG), \textit{Pantheon Exterior} (PE), \textit{Lincoln Memorial} (LM) and \textit{Palacio de Bellas Artes} (PBA). Each scene contains between 1000 and 1600 internet photos, a sparse point cloud, segmentation masks and a ground-truth mesh.

Following~\cite{sun2022neural}, we measure reconstruction quality by precision (P), recall (R), and F1 score, which reflect accuracy, completeness, and overall quality, respectively. Table~\ref{tab:overall} reports these metrics at three scene-specific distance thresholds, labeled Low, Medium, and High, which are listed in Table~\ref{tab:error}; Low is the strictest. Our ablation studies on Heritage-Recon use the Low threshold of each scene. Here and in the following, the mean over scenes is the unweighted average of the per-scene scores shown, rounded half up to the same precision.

\begin{figure*}[t]
\centering
    \includegraphics[width=\textwidth]{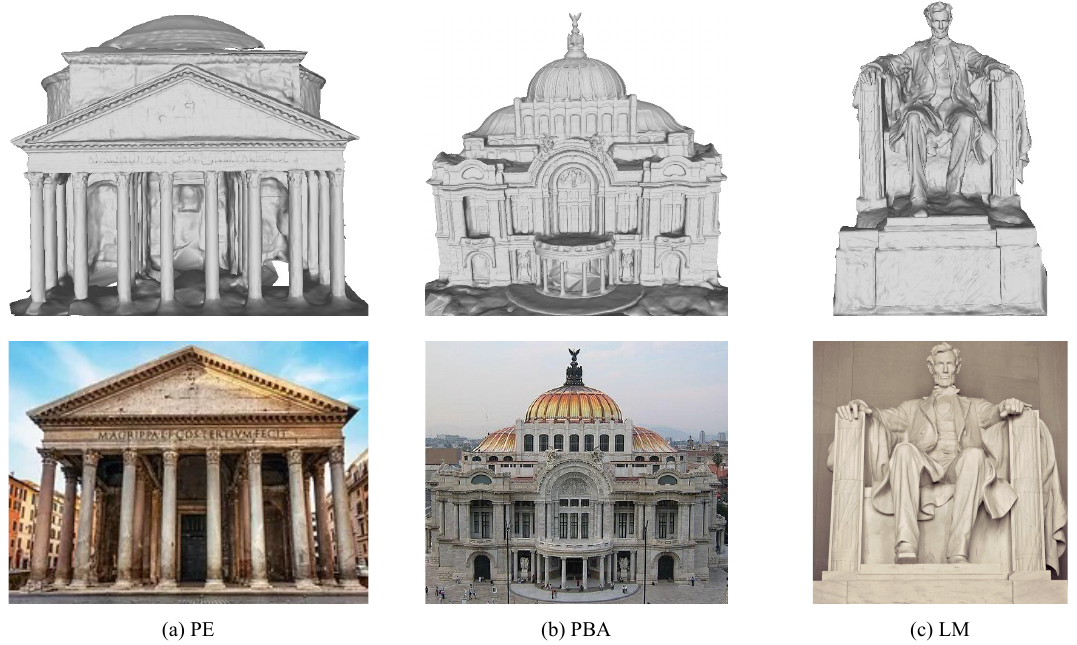}
  \caption{More qualitative results on the Heritage-Recon dataset~\cite{sun2022neural}: reconstructed 3D meshes of our method for (a) \textit{PE}, (b) \textit{PBA}, and (c) \textit{LM}, each shown above a reference photo. At the Low threshold, the F1 score increases from 49.6 for NeuralRecon-W to 54.7 for our method on \textit{PE}, from 66.4 to 67.5 on \textit{PBA}, and from 34.1 to 35.7 on \textit{LM} (Table~\ref{tab:overall}).}
   \label{fig:more}
\end{figure*}

For traditional methods, we compare with COLMAP~\cite{sch2016pw}, using two octree depths ($d=11$ and $13$) for its Poisson surface reconstruction. For learning-based approaches, we compare with NeuralRecon-W~\cite{sun2022neural}, which proposes a hybrid voxel- and surface-guided sampling technique to improve surface quality for in-the-wild scenes.

\begin{table}[t!]
\caption{Heritage-Recon evaluation thresholds~\cite{sun2022neural}. The thresholds are scene-specific distances in meters, and Low is the strictest.}
\centering
\begin{tabular}{l c c c c}
\toprule
\textbf{Level} & \textbf{BG} & \textbf{LM} & \textbf{PE} & \textbf{PBA}\\
\midrule
\textbf{Low}    &0.1 & 0.01 &0.04& 0.2 \\
\textbf{Medium} &0.2 &0.02 &0.06 &0.4\\
\textbf{High}   &0.3 & 0.03 &0.08 &0.6\\
\bottomrule
\end{tabular}
\label{tab:error}
\end{table}

We also compare with three further baselines: Neuralangelo~\cite{li2023neuralangelo}, which reconstructs high-fidelity surfaces for object-centric scenes and several large-scale scenes; Geo-NeuS~\cite{fu2022geo}, which directly supervises the SDF with sparse SfM points; and GeoMVSNet~\cite{zhang2023geomvsnet}, a recent learning-based MVS method. For GeoMVSNet, we reconstruct meshes from its dense point clouds using Poisson surface reconstruction~\cite{kazhdan2006poisson}.  Table~\ref{tab:overall} also includes VisMVSNet~\cite{zhang2020visibility} and NeRF-W~\cite{martin2021nerf}. The results of COLMAP, VisMVSNet, NeRF-W, and NeuralRecon-W are taken from~\cite{sun2022neural}.

As shown in Table~\ref{tab:overall}, our method achieves the highest mean F1 score among the compared methods at all three thresholds, improving on NeuralRecon-W by 2.6, 1.4, and 2.1 points at the Low, Medium, and High thresholds, respectively. At the scene level, it achieves the highest F1 score on all scenes except PE, where COLMAP ($d=13$) scores higher.
Figs.~\ref{fig:bg} and~\ref{fig:baseline_bg_fig} compare the reconstructions of BG. In the close-up views, our method recovers finer details than NeuralRecon-W, while the other baselines produce noisier or less complete surfaces in these regions, consistent with the scene-level F1 scores in Table~\ref{tab:overall}. Fig.~\ref{fig:more} shows our reconstructions of the other three scenes.

\subsection{Evaluation on Tanks and Temples Dataset} 

We also assessed reconstruction quality on the widely used Tanks and Temples dataset~\cite{Knapitsch2017} including large-scale indoor/outdoor scenes. Each scene comprises between 263 and 1107 images taken with a handheld monocular RGB camera, with ground-truth data acquired via a LiDAR sensor. Following prior work~\cite{li2023neuralangelo}, \textit{F1 score} is used for surface evaluation on the six scenes listed in Table~\ref{tab:tnt}. The results of COLMAP and Neuralangelo are taken from~\cite{li2023neuralangelo}, and we run GeoMVSNet and Geo-NeuS following the parameter settings described in their papers.

\begin{figure*}[t]
\centering
\includegraphics[width=\textwidth]{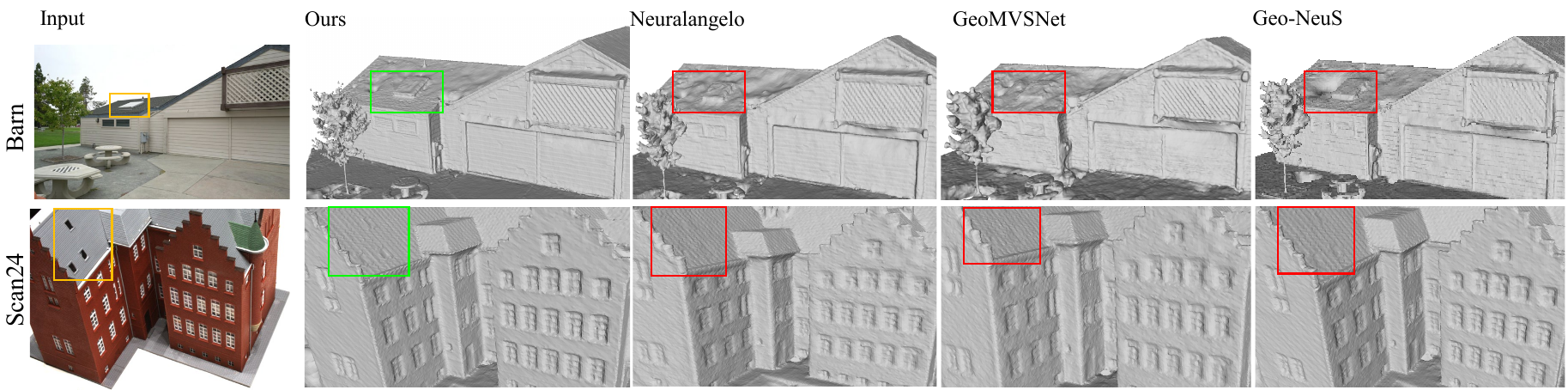}
\caption{Qualitative comparison on \textit{Barn} from Tanks and Temples and \textit{Scan24} from DTU. The highlighted regions are outlined in green for our method and in red for the baselines. On \textit{Barn}, the F1 scores (higher is better) are 0.71 for our method, 0.70 for Neuralangelo, 0.58 for GeoMVSNet, and 0.31 for Geo-NeuS (Table~\ref{tab:tnt}). On \textit{Scan24}, the Chamfer distances (lower is better) are 0.32, 0.37, 0.36, and 0.38, respectively (Table~\ref{tab:dtu}).}
\label{fig:baseline_tnt_fig}
\end{figure*}

\begin{table}[t]
    \centering
    \caption{Performance comparison on the Tanks and Temples dataset (F1$\uparrow$). The best results are in \textbf{bold}, and the second-best are \underline{underlined}.}
    \begingroup
    \fontsize{8}{10}\selectfont
    \renewcommand{\arraystretch}{1.0}
    \setlength{\tabcolsep}{0.5pt}
    \begin{adjustbox}{max width=\columnwidth}
    \begin{tabular}{@{}lccccc@{}}
        \toprule
         & \footnotesize{\shortstack{COLMAP\\\cite{sch2016pw}}} & {\footnotesize{\shortstack{GeoMVSNet\\\cite{zhang2023geomvsnet}}}} & { \footnotesize{\shortstack{Geo-NeuS\\\cite{fu2022geo}}}} & { \footnotesize{\shortstack{Neuralangelo\\\cite{li2023neuralangelo}}}} & {\footnotesize{Ours}} \\
        \midrule 
        \footnotesize{Barn}         &0.55  &0.58   &0.31   &\underline{0.70}  &\textbf{0.71} \\
        \footnotesize{Caterpillar}  &0.01  &\underline{0.35}   &0.30   &\textbf{0.36}  &0.34 \\
        \footnotesize{Courthouse}   &0.11  &0.21   &0.19   &\underline{0.28}  &\textbf{0.29} \\
        \footnotesize{Ignatius}     &0.22  &0.84   &0.88   &\underline{0.89}  &\textbf{0.90} \\
        \footnotesize{Meetingroom}  &0.19  &\underline{0.30}   &0.27   &\textbf{0.32}  &\underline{0.30} \\
        \footnotesize{Truck}       &0.19  &0.45   &0.43   &\underline{0.48}  &\textbf{0.50} \\
        \midrule
        \footnotesize{\textbf{Mean}} &0.21 &0.46  &0.40   &\textbf{0.51} &\textbf{0.51} \\
        \bottomrule
    \end{tabular}
    \end{adjustbox}
    \endgroup
    \label{tab:tnt}
\end{table}

As shown in Table~\ref{tab:tnt}, our method and Neuralangelo achieve the highest mean F1 score ($0.51$). At the scene level, our method performs best on four of the six scenes, and Neuralangelo on the other two, \textit{Caterpillar} and \textit{Meetingroom}. On \textit{Caterpillar}, GeoMVSNet also scores slightly higher than ours. A possible reason for our lower score on this scene is that it contains many thin and partially occluded structures, where SfM coverage may be limited and filtering may remove many normal priors, leaving less geometric guidance. Fig.~\ref{fig:baseline_tnt_fig} shows that our method recovers the skylight on the roof of \textit{Barn} more clearly than the other methods.

\begin{table}[t]
    \centering
    %\caption{Quantitative comparison of Chamfer distance~($\downarrow$) with different baselines on the DTU dataset without mask.}
    \caption{Quantitative comparison of Chamfer distance ($\downarrow$) with different baselines on the DTU dataset. None of the methods uses object masks for reconstruction. The best mean is in \textbf{bold}, and the second-best is \underline{underlined}.}
    \begingroup
    \fontsize{8}{10}\selectfont
    \renewcommand{\arraystretch}{1.0}
    \setlength{\tabcolsep}{0.8pt}
    \begin{tabular}{@{}cccccc@{}}
        \toprule
        { \textbf{\footnotesize{ScanID}}} & \footnotesize{\shortstack{COLMAP\\\cite{sch2016pw}}} & {\footnotesize{\shortstack{GeoMVSNet\\\cite{zhang2023geomvsnet}}}} & { \footnotesize{\shortstack{Geo-NeuS\\\cite{fu2022geo}}}} & { \footnotesize{\shortstack{Neuralangelo\\\cite{li2023neuralangelo}}}} &{\footnotesize{Ours}} \\
        \midrule
             24  &0.45   &0.36   &0.38   &0.37   &0.32 \\
        37  &0.91   &0.52   &0.54   &0.72   &0.52 \\
        40  &0.37   &0.35   &0.34   &0.35   &0.31 \\
        55  &0.37   &0.35   &0.36   &0.35   &0.34 \\
        63  &0.90   &0.86   &0.80   &0.87   &0.78 \\
        65  &1.00   &0.48   &0.45   &0.54   &0.44 \\
        69  &0.54   &0.34   &0.41   &0.53   &0.36 \\
        83  &1.22   &0.98   &1.03   &1.29   &0.90 \\
        97  &1.08   &0.79   &0.84   &0.97   &0.79 \\
        105 &0.64   &0.54   &0.55   &0.73   &0.57 \\
        106 &0.48   &0.43   &0.46   &0.47   &0.44 \\
        110 &0.59   &0.43   &0.47   &0.74   &0.43 \\
        114 &0.32   &0.30   &0.29   &0.32   &0.33 \\
        118 &0.45   &0.31   &0.36   &0.41   &0.31 \\
        122 &0.43   &0.33   &0.35   &0.43   &0.32 \\
        \midrule
        { \textbf{Mean}} &0.65 &\underline{0.49} &0.51 &0.61 &\textbf{0.48} \\
        \bottomrule
    \end{tabular}
    \endgroup
    \label{tab:dtu}
\end{table}

\subsection{Evaluation on DTU Dataset}
Although the proposed method targets in-the-wild scenes with foreground occlusions and varying appearances, we also evaluate it on the DTU dataset~\cite{jensen2014large}, which was captured in a controlled indoor environment. Results are shown in Fig.~\ref{fig:baseline_tnt_fig} and Table~\ref{tab:dtu}. Reconstruction quality is measured using \textit{Chamfer distance}, following NeuS~\cite{wang2021neus}. None of the methods uses object masks for reconstruction. The results of COLMAP and Geo-NeuS are taken from~\cite{fu2022geo} and those of Neuralangelo from~\cite{li2023neuralangelo}, and we run GeoMVSNet ourselves in the same setting.

Our method achieves the lowest mean Chamfer distance ($0.48$), closely followed by GeoMVSNet ($0.49$). It thus performs comparably to the strongest baseline in this controlled setting, while its intended application of reconstruction from internet photos is evaluated on Heritage-Recon.

\begin{figure}[t]
\centering
\includegraphics[width=\columnwidth]{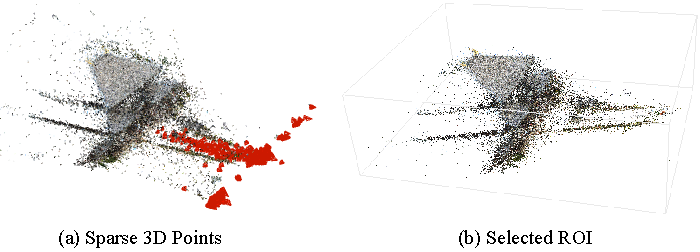}
    \caption{ Sparse 3D points and selected reconstruction region shown in the bounding box for \textit{Todai-ji Temple}.
  }
   \label{fig:bbox}

\end{figure}

\begin{figure*}[t]
\centering
    \includegraphics[width=\textwidth]{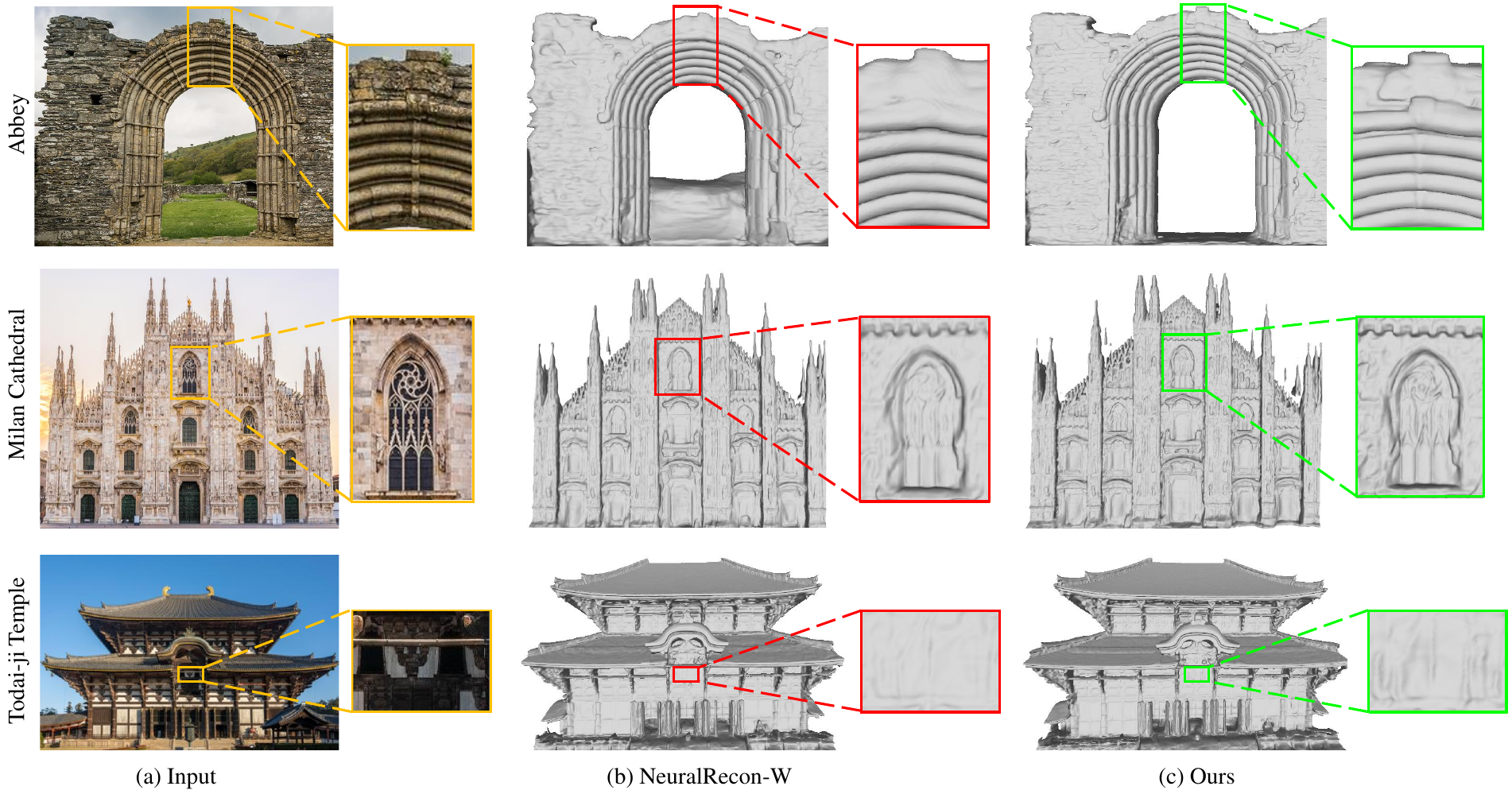}
  \caption{Qualitative comparison with NeuralRecon-W~\cite{sun2022neural} on three additional landmarks: \textit{Strata Florida Abbey}, with photos collected from Flickr, and \textit{Milan Cathedral} and \textit{Todai-ji Temple}, with photos from the Image Matching Challenge dataset~\cite{Jin2020}. The close-ups, outlined in red for NeuralRecon-W and in green for our method, show that our method recovers finer details.}
   \label{fig:custom}
\end{figure*}

\subsection{Reconstruction on Custom Scenes}

To further validate the effectiveness of the proposed method, we collected unstructured photos of three landmarks, \textit{Strata Florida Abbey} (151 photos) from Flickr, and \textit{Milan Cathedral} (120 photos) and \textit{Todai-ji Temple} (900 photos) from the Image Matching Challenge dataset~\cite{Jin2020}. Following the same preprocessing procedure as~\cite{sun2022neural}, we first run COLMAP to generate sparse points for each scene and manually select the reconstruction regions, as the sparse points are often noisy and make it difficult to determine these regions automatically (see Fig.~\ref{fig:bbox} for an example). We generate semantic segmentation masks and normal priors as explained in Sec.~\ref{sec:method}. Finally, we train our model on these landmarks and reconstruct their 3D surfaces.
 
Since these scenes have no ground-truth geometry, we compare our method with NeuralRecon-W~\cite{sun2022neural} qualitatively in Fig.~\ref{fig:custom}. In the enlarged regions, our reconstructions show finer details, such as the stone blocks above the arch of \textit{Strata Florida Abbey} and the window tracery of \textit{Milan Cathedral}.

\begin{figure*}[t]
\centering
    \includegraphics[width=\textwidth]{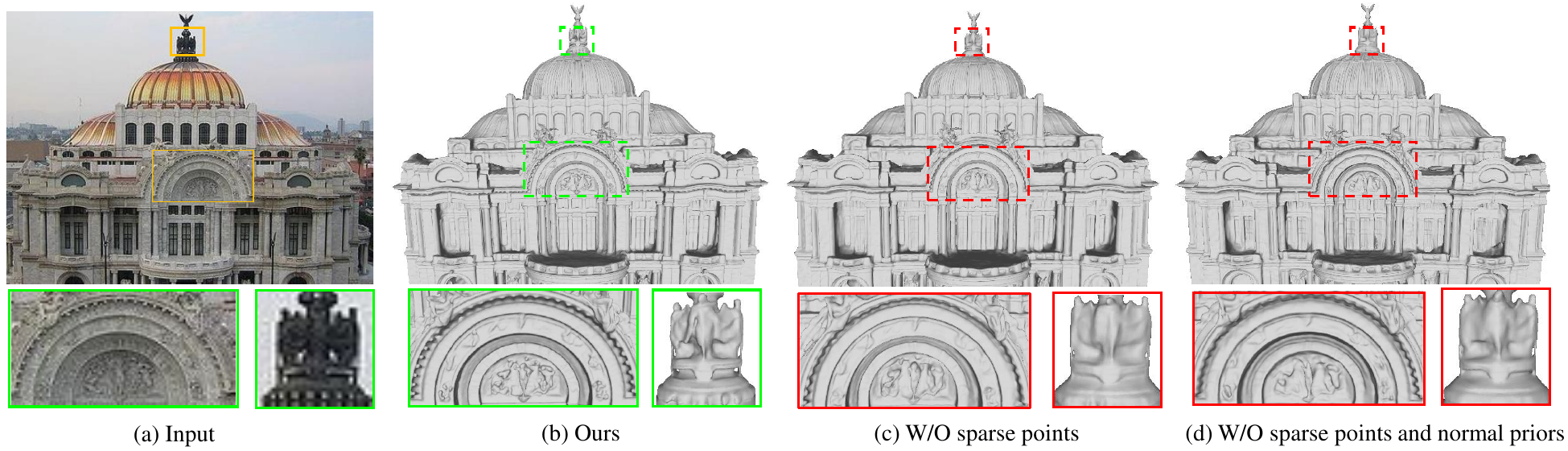}
  \caption{Ablation of the geometric priors on \textit{PBA}. (a) Input image. (b) Our full method. (c) Without sparse points. (d) Without sparse points and normal priors, i.e., the NeuralRecon-W baseline. The close-ups show the relief in the central arch and the sculpture on top of the dome, outlined in green for our full method and in red for the other two settings.}
   \label{fig:abla}
\end{figure*}

\subsection{Ablation Studies}

\subsubsection{Ablation of Different Priors}

We performed an ablation analysis to evaluate the effectiveness of our sparse point priors and normal priors for the reconstruction. We used NeuralRecon-W~\cite{sun2022neural}, on which our method is built, as the baseline. We then considered three variants of geometric priors in our method: (1) using the sparse point priors only; (2) using the normal priors only; (3) using both priors as done in our full method. We evaluated the reconstruction quality of these approaches on the \textit{BG} and \textit{PE} scenes from Heritage-Recon~\cite{sun2022neural} using the Low error threshold.   For each setting on \textit{BG}, we report the mean and sample standard deviation over five runs, which were performed separately from the single runs reported in the other tables. F1 is computed from the paired precision and recall within each run before averaging.

\begin{table}[t]
\caption{Ablation of different priors on the \textit{BG} scene from Heritage-Recon~\cite{sun2022neural}. The best results are in \textbf{bold}. Each entry is the mean $\pm$ sample standard deviation over five new runs, evaluated under the Low threshold; these runs are separate from the single runs reported in the other tables. The first row is the NeuralRecon-W baseline from these runs, so it differs from the result reported in~\cite{sun2022neural} and listed in Table~\ref{tab:overall}.}
\centering%把表居中
\begingroup
\fontsize{8}{10}\selectfont
\renewcommand{\arraystretch}{1.15}
\setlength{\tabcolsep}{2.5pt}
\resizebox{\columnwidth}{!}{%
%\begin{tabular}{c|c|c|c|c|c}
\begin{tabular}{c|c|c|c|c}
\hline %{ Baseline} & 
{\shortstack{Sparse\\points}} & {\shortstack{Normal\\priors}} &{ \textbf{P}$\uparrow$} & { \textbf{R}$\uparrow$} & { \textbf{F1}$\uparrow$} \\
\hline
 %$\checkmark$ & 
 $\times$ & $\times$ & 61.76$\pm$0.25 & 47.37$\pm$0.32 & 53.62$\pm$0.26 \\
 %$\checkmark$ & 
 $\checkmark$ & $\times$ & 64.60$\pm$0.47 & 50.11$\pm$0.38 & 56.44$\pm$0.27 \\
 %$\checkmark$ & 
 $\times$ & $\checkmark$ & 63.17$\pm$0.14 & 49.45$\pm$0.37 & 55.47$\pm$0.19 \\
 %$\checkmark$ & 
 $\checkmark$ & $\checkmark$ & \textbf{65.80$\boldsymbol{\pm}$0.45} & \textbf{51.11$\boldsymbol{\pm}$0.38} & \textbf{57.53$\boldsymbol{\pm}$0.11} \\
\hline
\end{tabular}
}
\endgroup
\label{tab:abla}
\end{table}

\begin{table}[t]
\caption{Ablation of different priors on the \textit{PE} scene from Heritage-Recon~\cite{sun2022neural}. 
The best results are in \textbf{bold}. Each entry is from a single run, evaluated under the Low threshold. The baseline and full-model results are those in Table~\ref{tab:overall}.
}
\centering
\begingroup
\fontsize{8}{10}\selectfont
\renewcommand{\arraystretch}{1.15}
\setlength{\tabcolsep}{4pt}
%\begin{tabular}{c|c|c|c|c|c}
\begin{tabular}{c|c|c|c|c}
\hline 
%{Baseline} & 
{\shortstack{Sparse\\points}} & {\shortstack{Normal\\priors}} & {\textbf{P}$\uparrow$} & {\textbf{R}$\uparrow$} & {\textbf{F1}$\uparrow$} \\
\hline
%$\checkmark$ & 
$\times$ & $\times$ & 48.5 & 50.8 & 49.6 \\
%$\checkmark$ & 
$\checkmark$ & $\times$ & $52.6$ & $53.2$ & 52.9 \\
%$\checkmark$ & 
$\times$ & $\checkmark$ & $50.4$ & $51.3$ & 50.8 \\
%$\checkmark$ & 
$\checkmark$ & $\checkmark$ & \textbf{55.6} & \textbf{53.8} & \textbf{54.7} \\
\hline
\end{tabular}
\endgroup
\label{tab:abla_pe}
\end{table}

As shown in Tables~\ref{tab:abla} and~\ref{tab:abla_pe}, both sparse-point and normal priors improve reconstruction, and their combination achieves the highest F1 scores on \textit{BG} and \textit{PE}. Sparse point priors yield larger gains than normal priors in both scenes. Combining the two priors improves the mean F1 score by 3.91 points on \textit{BG} and the single-run F1 score by 5.1 points on \textit{PE}.

We also provide qualitative visualizations on \textit{PBA} in Fig.~\ref{fig:abla}. After removing the sparse 3D points, the quality of the reconstructed mesh degrades in the local details shown in the green and red rectangles, namely the relief in the central arch and the sculpture on top of the dome. If normal priors are further removed, then more details are lost, and the reconstruction quality is further degraded. These quantitative and qualitative results show the effectiveness of both priors in high-quality reconstruction.

\subsubsection{Sensitivity of Loss Weights}
\label{sec:lambda_sens}
We also analyzed the sensitivity of our method to the loss weights defined in Sec.~\ref{sec:loss}. With the weight $\lambda_1$ of the color loss fixed, we varied the weights $\lambda_2$ to $\lambda_5$ of the normal, displacement compensation, Eikonal, and mask losses one at a time, keeping the other weights at their default values. Specifically, we halved and doubled $\lambda_2$ and $\lambda_3$, and decreased and increased $\lambda_4$ and $\lambda_5$ by a factor of 10. We evaluated the reconstruction quality on the \textit{BG} scene using the Low error threshold. As shown in Table~\ref{tab:lambda}, the default value of each weight gives the highest F1 score (56.8), and the other values lower it by 0.3 to 0.9 points, with the largest decreases caused by increasing $\lambda_4$ and $\lambda_5$. Even the lowest score (55.9) is higher than the mean F1 score of the baseline in Table~\ref{tab:abla} (53.62), so the improvement from our priors does not depend on a precise choice of these weights.

\begin{table}[t]
\caption{Sensitivity of loss weights on the \textit{BG} scene under the Low threshold (F1$\uparrow$). Each group varies one weight with the others fixed at their default values, which are shown in \textbf{bold}. Each entry is from a single run.}%标题
\centering
\setlength{\tabcolsep}{3.5pt}
\begin{tabular}{cccccccc}
\toprule
\multicolumn{2}{c}{$\lambda_2$} & \multicolumn{2}{c}{$\lambda_3$} & \multicolumn{2}{c}{$\lambda_4$} & \multicolumn{2}{c}{$\lambda_5$} \\
%\midrule
\cmidrule(lr){1-2} \cmidrule(lr){3-4} \cmidrule(lr){5-6} \cmidrule(lr){7-8}
{Value} & {F1$\uparrow$} & {Value} & {F1$\uparrow$} & {Value} & {F1$\uparrow$} & {Value} & {F1$\uparrow$} \\
\midrule
$0.5$ & 56.1 & $0.5$ & 56.3 & $0.001$ & 56.5 & $0.01$ & 56.5 \\
$\mathbf{1.0}$ & $\mathbf{56.8}$ & $\mathbf{1.0}$ & $\mathbf{56.8}$ & $\mathbf{0.01}$ & $\mathbf{56.8}$ & $\mathbf{0.1}$ & $\mathbf{56.8}$ \\
$2.0$ & 56.2 & $2.0$ & 56.4 & $0.1$ & 55.9 & $1.0$ & 56.0 \\
\bottomrule
\end{tabular}
\label{tab:lambda}
\end{table}

\subsubsection{Displacement Compensation}\label{sec:abl_comp}
The original sparse points used for explicit supervision may contain noise and can cause incorrect prediction of the SDF value. Displacement compensation (Sec.~\ref{sec:point_comp}) is designed to reduce the effect of such noise. To isolate its effect, we compare it with raw sparse point supervision that penalizes the SDF values directly at the SfM points, on the same NeuralRecon-W backbone without normal priors (Table~\ref{tab:bias}). 
Displacement compensation improves the F1 score over raw supervision by 1.0 point on \textit{BG} and 1.4 points on \textit{PE}. Table~\ref{tab:bias} also lists Geo-NeuS~\cite{fu2022geo}, which likewise supervises the SDF directly at the SfM points but is built on NeuS~\cite{wang2021neus}; its F1 scores are lower than those of all three settings on both scenes.

To further evaluate its robustness to noise, we added isotropic Gaussian noise to the SfM points of \textit{BG} and trained both settings with the perturbed points (Table~\ref{tab:noise}). The noise has a standard deviation $\sigma$ along each coordinate axis. The largest value of $\sigma$ is $0.10$\,m, which equals the Low threshold of this scene. 
As $\sigma$ increases from 0 to $0.10$\,m, the F1 score decreases by 3.4 points with raw supervision but only by 1.8 points with compensation, and the advantage of compensation grows from 1.0 to 2.6 points. These results show that displacement compensation makes the reconstruction less sensitive to Gaussian noise in the SfM points.

\begin{table}[t]
\caption{Ablation of displacement compensation on the \textit{BG} and \textit{PE} scenes under the Low threshold. The best results are in \textbf{bold}. The baseline results are those of NeuralRecon-W in Table~\ref{tab:overall}.}
\centering
\begingroup
\fontsize{8}{10}\selectfont
\setlength{\tabcolsep}{0.5pt}
\renewcommand{\arraystretch}{1.1}
\begin{tabular*}{\columnwidth}{@{\extracolsep{\fill}}lcccccc@{}}
\toprule
& \multicolumn{3}{c}{\textit{BG}} & \multicolumn{3}{c}{\textit{PE}} \\
\cmidrule(lr){2-4} \cmidrule(lr){5-7}
\textbf{Method} & P$\uparrow$ & R$\uparrow$ & F1$\uparrow$ & P$\uparrow$ & R$\uparrow$ & F1$\uparrow$ \\
\midrule
Baseline & $62.6$ & $47.9$ & $54.3$ & 48.5 & 50.8 & 49.6 \\
\shortstack[l]{+ Raw sparse points} & $63.4$ & $48.3$ & $54.8$ & $50.3$ & $52.7$ & 51.5 \\
+ Compensation & $\mathbf{64.2}$ & $\mathbf{49.3}$ & $\mathbf{55.8}$ & \textbf{52.6} & \textbf{53.2} & \textbf{52.9} \\
\midrule
Geo-NeuS & $62.9$ & $45.4$ & $52.8$ & 46.3 & 48.2 & 47.2 \\
\bottomrule
\end{tabular*}
\endgroup
\label{tab:bias}
\end{table}

\begin{table}[t]
\caption{Effect of Gaussian noise added to the SfM points on the \textit{BG} scene under the Low threshold (F1$\uparrow$). The noise has a standard deviation $\sigma$ along each coordinate axis. The best results are in \textbf{bold}.}%标题
\centering
\setlength{\tabcolsep}{5pt}
\begin{tabular}{l c c}
\toprule
{ $\sigma$ (m)} & {+ Raw sparse points} & {+ Compensation} \\
\midrule
{ $0$ (original)} & $54.8$ & $\mathbf{55.8}$ \\
{ $0.02$} & $54.3$ & $\mathbf{55.5}$ \\
{ $0.05$} & $53.1$ & $\mathbf{54.9}$ \\
{ $0.10$} & $51.4$ & $\mathbf{54.0}$ \\
\bottomrule
\end{tabular}
\label{tab:noise}
\end{table}

\subsubsection{Normal Prior Filtering Strategy} We leverage the edge prior and the reprojection consistency check to filter unreliable normal priors. To verify the effectiveness of this strategy, we conducted an ablation study and the corresponding results are reported in Table~\ref{tab:normal}.
Except for the baseline, all settings use normal priors; none of the settings uses sparse point priors. Table~\ref{tab:normal} shows that unfiltered normal priors do not improve on the baseline on either scene. Edge filtering improves on them, and reprojection consistency checking provides a further improvement. Together, the two filters improve F1 over unfiltered normals by 1.7 points on \textit{BG} and 1.9 points on \textit{PE}, despite retaining fewer normal priors.

\begin{table}[t]
\caption{Ablation of our normal prior filtering strategy on the \textit{BG} and \textit{PE} scenes under the Low threshold. The best results are in \textbf{bold}. Retained is the percentage of normal priors kept after filtering. The baseline results are those of NeuralRecon-W in Table~\ref{tab:overall}.}
\centering
\setlength{\tabcolsep}{3pt}
\setlength{\aboverulesep}{0pt}
\setlength{\belowrulesep}{0pt}
\renewcommand{\arraystretch}{1.3}
\begin{tabular*}{\columnwidth}{@{\extracolsep{\fill}}llcccc@{}}
\toprule
 & \textbf{Method} & Retained & P$\uparrow$ & R$\uparrow$ & F1$\uparrow$ \\
\midrule
\multirow{4}{*}{\textit{BG}}
 & Baseline             & 0\% & 62.6 & 47.9 & 54.3 \\
 & + Raw normal priors  & 100\%     & 62.1 & 47.1 & 53.6 \\
 & + Edge filtering           & 63\%      & 62.9 & 48.1 & 54.5 \\
 & + Reprojection check & 34\%      & \textbf{63.4} & \textbf{49.0} & \textbf{55.3} \\
\midrule
\multirow{4}{*}{\textit{PE}}
 & Baseline             & 0\%   & 48.5 & 50.8 & 49.6 \\
 & + Raw normal priors  & 100\% & 48.2 & 49.6 & 48.9 \\
 & + Edge filtering     & 68\%  & 49.8 & 51.0 & 50.4 \\
 & + Reprojection check & 39\%  & \textbf{50.4} & \textbf{51.3} & \textbf{50.8} \\
\bottomrule
\end{tabular*}
\label{tab:normal}
\end{table}

\subsubsection{Sensitivity of the Reprojection Threshold}
\label{sec:reproj_sens}
We also evaluated how the threshold $\tau$ of the reprojection consistency check affects the reconstruction, using $\tau = 0.5$, $1$, and $2$ pixels. The other settings are the same as in the \textit{BG} `+ Reprojection check' row of Table~\ref{tab:normal}: we used normal priors with edge filtering and without sparse points, and evaluated the reconstruction quality on the \textit{BG} scene using the Low error threshold. As shown in Table~\ref{tab:reproj}, increasing $\tau$ from 0.5 to 2 pixels raises the percentage of retained normal priors from 22\% to 47\%. The default threshold of 1 pixel gives the highest precision, recall, and F1 score, while the other two thresholds lower the F1 score by 0.5 and 0.3 points, respectively. This is consistent with a trade-off between the amount and the reliability of the normal priors: a smaller threshold leaves fewer priors for supervision, whereas a larger one retains more priors that may be unreliable.
\begin{table}[t]
\caption{Sensitivity of the reprojection threshold $\tau$ on the \textit{BG} scene under the Low threshold. The best results are in \textbf{bold}. Retained is the percentage of normal priors kept after filtering.}%标题
\centering
\setlength{\tabcolsep}{4pt}
\begin{tabular}{l c c c c}
\toprule
{ $\tau$ (pixels)} & {Retained} & {P$\uparrow$} & {R$\uparrow$} & {F1$\uparrow$} \\
\midrule
{ $0.5$} & 22\% & 63.0 & 48.5 & 54.8 \\
{ $1.0$ (default)} & 34\% & \textbf{63.4} & \textbf{49.0} & \textbf{55.3} \\
{ $2.0$} & 47\% & 63.1 & 48.7 & 55.0 \\
\bottomrule
\end{tabular}
\label{tab:reproj}
\end{table}

\begin{figure*}[t]
   \centering
   \includegraphics[width=\linewidth]{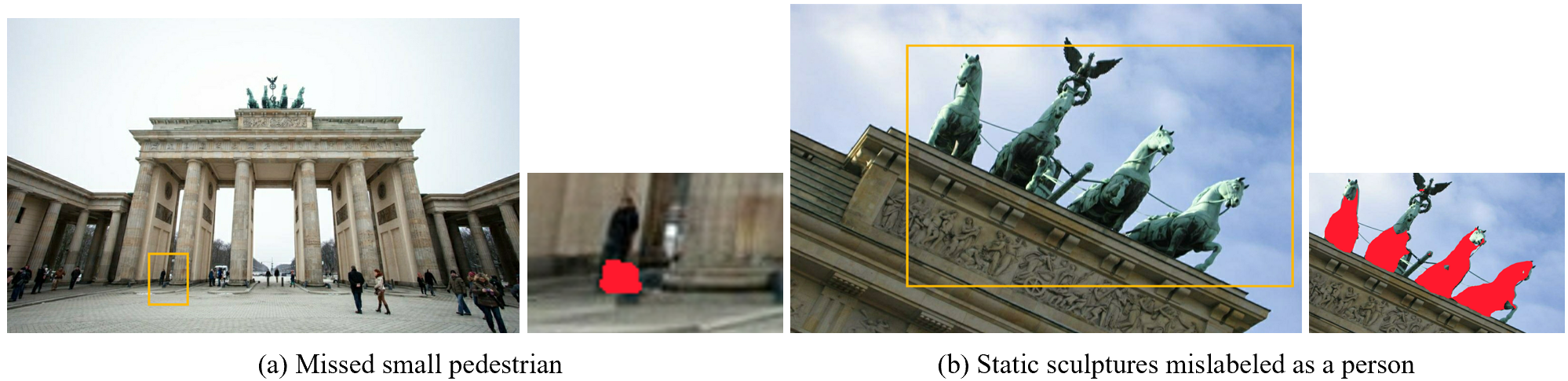}
   \caption{Examples of segmentation errors on \textit{BG}. The yellow box marks the region enlarged on the right, where the pixels labeled as a person are shown in red. (a) Most of a small pedestrian is missed. (b) Parts of the static sculptures are labeled as a person.}
   \label{fig:seg_fail}
\end{figure*}

\subsection{Computational Cost}\label{sec:cost}

\begin{table}[!t]
\caption{Computational cost on \textit{BG} (four A100 GPUs, 20 epochs; training memory per GPU). Training totals sum baseline and increments.}
\centering
%\begin{adjustbox}{max width=\columnwidth}
\begingroup
\fontsize{8}{10}\selectfont
\setlength{\tabcolsep}{1pt}
\renewcommand{\arraystretch}{1.05}

\begin{tabular*}{\columnwidth}{@{\extracolsep{\fill}}llcc@{}}
\toprule
{\footnotesize{{Stage}}} &{\footnotesize{{Method/Component}}} & \footnotesize{\shortstack{{Time}\\(h)}} & \footnotesize{\shortstack{{Memory}\\(GB)}} \\
\midrule
\multirow{2}{*}[1pt]{\footnotesize{Preprocessing}} & Normal prediction & $0.47$ & $4.10$ \\
              & Edge detection & $0.10$ & $2.30$ \\
\midrule
\multirow{4}{*}[1pt]{\footnotesize{Training}} & \shortstack[l]{Baseline (NeuralRecon-W)} & $8.50$ & $16.20$ \\
         & +\hspace{1pt}{Displacement compensation} & $+0.99$ & $+0.85$ \\
         & +\hspace{1pt}Consistency check & $+0.70$ & $+0.93$ \\
         & \textbf{Ours (total)} & $\mathbf{10.19}$ & $\mathbf{17.98}$ \\
\midrule
\multirow{2}{*}[1pt]{\footnotesize{Inference}} & Baseline (mesh) & 0.22 & 6.23 \\
          & Ours (mesh) & 0.26 & 6.56\\
\bottomrule
\end{tabular*}

\endgroup
%\end{adjustbox}
\par\smallskip
\noindent\parbox{\columnwidth}{\fontsize{8}{10}\selectfont\textbf{Inference = mesh extraction:} batched SDF queries $\rightarrow$ Marching Cubes $\rightarrow$ mesh export (both methods).}
\label{tab:cost}
\end{table}

Table~\ref{tab:cost} compares the computational cost on \textit{BG} for 20 epochs on four NVIDIA A100 GPUs. The NeuralRecon-W baseline requires $8.50$\,h and $16.20$\,GB per GPU. Adding the listed training increments gives calculated totals of $10.19$\,h and $17.98$\,GB per GPU, increases of $1.69$\,h ($19.9\%$) and $1.78$\,GB ($11.0\%$), respectively. Displacement compensation contributes 0.99\,h and 0.85\,GB. The reprojection consistency check is performed during training using the current SDF, with increments of $0.70$\,h and $0.93$\,GB.

Normal prediction and edge detection are performed once before training and cached. They require approximately $0.47$\,h/$4.10$\,GB and $0.10$\,h/$2.30$\,GB, respectively. Their listed times sum to $0.57$\,h when run sequentially, excluding shared data preparation and reported separately from training. These preprocessing memory figures are stage-specific and are not summed.

Reconstruction inference here means mesh extraction: batched queries of the trained SDF, followed by Marching Cubes and mesh export. Mesh extraction takes $0.22$\,h and $6.23$\,GB for NeuralRecon-W, compared with $0.26$\,h and $6.56$\,GB for our method. Both methods use the same geometry-network architecture; extraction requires only $f_{geo}$, without the prior predictors or consistency checking.

\section{Conclusion and Limitations}\label{sec5}

In this paper, we proposed a geometric prior-guided neural implicit surface reconstruction method for unstructured internet photos. To accurately recover finer geometric details, we introduced sparse 3D points to regularize neural SDF learning and proposed a displacement compensation strategy that evaluates the SDF at points projected toward the current surface, which reduces the effect of noise in these points. We additionally utilized robust normal priors filtered by edge detection and a reprojection consistency check to further improve reconstruction quality. Experiments show that our method achieves the best average performance among the compared methods on Heritage-Recon and performs comparably to the strongest baselines on Tanks and Temples and DTU. Our ablations show that each component improves the reconstruction, and that displacement compensation reduces its sensitivity to Gaussian noise in the SfM points.
 
Although the proposed method can boost surface quality, its calculated training time is 19.9\% longer than that of NeuralRecon-W, and it requires additional preprocessing (Sec.~\ref{sec:cost}). Additionally, inaccurate pose estimation from SfM can affect reconstruction accuracy. Moreover, the semantic masks used to exclude transient objects (Sec.~\ref{sec:preliminary}) may be inaccurate (see Fig.~\ref{fig:seg_fail}). If a small or partially occluded pedestrian or vehicle is missed, its pixels remain in the color loss and may produce floating artifacts. Conversely, if permanent structures such as sculptures are labeled as transient objects, their observations are excluded from training, which may lead to incomplete reconstruction. In future work, we will explore recent Gaussian Splatting techniques~\cite{kerbl20233d} to impose stronger geometric constraints on the scene surfaces and better handle dynamic objects for in-the-wild surface reconstruction.

\section*{Declarations}

\bmhead{Author Contribution} Lintao Xiang was responsible for conceptualization, methodology, code implementation, writing the original draft, and reviewing and editing the manuscript. Hongpei Zheng contributed to code implementation, the formal analysis and visualization. Bailin Deng was responsible for conceptualization, methodology, supervision, reviewing and revising the manuscript. Hujun Yin was responsible for conceptualization, methodology, supervision, reviewing and revising the manuscript.

\bmhead{Data Availability Statement} The Heritage-Recon~\cite{sun2022neural}, Tanks and Temples~\cite{Knapitsch2017}, DTU~\cite{jensen2014large}, and Image Matching Challenge~\cite{Jin2020} datasets used in this study are publicly available from their original sources.

\bmhead{Funding} No funding was received to assist with the preparation of this manuscript.

\bmhead{Competing interests} The authors declare no competing interests.

% Keep the reference list readable without an isolated final entry.
\setlength{\bibsep}{6pt}
%\balance
\bibliography{sn-bibliography}% common bib file


\end{document}